\documentclass[11pt]{article}

\usepackage{acl}

\usepackage{times}
\usepackage{latexsym}
\usepackage{graphicx}
\usepackage{booktabs}
\usepackage{subcaption}
\usepackage{newfloat}
\usepackage{listings}
\usepackage{array}
\usepackage{graphicx}   
\usepackage{booktabs}    
\usepackage{multirow}
\usepackage{makecell}
\DeclareCaptionStyle{ruled}{labelfont=normalfont,labelsep=colon,strut=off} %
\usepackage{amsmath}
\usepackage[T1]{fontenc}
\usepackage{xcolor}         
\usepackage{colortbl}       
\usepackage[utf8]{inputenc}
\usepackage{booktabs}
\usepackage{xcolor}
\usepackage{colortbl}
\usepackage{graphicx}
\usepackage{booktabs}
\usepackage{multirow}
\usepackage{xcolor}
\usepackage{pifont}
\usepackage{fontawesome}
\usepackage{amssymb}
\usepackage{float}
\usepackage{placeins}
\usepackage{enumitem}
\usepackage{diagbox} 
\usepackage{tikz}
\definecolor{best}{RGB}{255,200,200}      
\definecolor{secondbest}{RGB}{255,230,230} 

\usepackage{makecell}

\newcommand{\best}[1]{\cellcolor{best}\textbf{#1}}
\newcommand{\secondbest}[1]{\cellcolor{secondbest}#1}

\usepackage{microtype}

\usepackage{inconsolata}

\usepackage{graphicx}

\title{Perception, Understanding and Reasoning: \\ A Multimodal Benchmark for Video Fake News Detection}

\author{
 \textbf{Yakun Cui\textsuperscript{1}},
 \textbf{Peng Qi\textsuperscript{2}},
 \textbf{Fushuo Huo\textsuperscript{3}},
 \textbf{Weijie Shi\textsuperscript{1}},
 \textbf{Juntao Dai\textsuperscript{5}},
\\
 \textbf{Hang Du\textsuperscript{4}},
 \textbf{Zhenghao Zhu\textsuperscript{1}},
 \textbf{Sirui Han\textsuperscript{1}}
 \textbf{Yike Guo\textsuperscript{1}}
\\
 \textsuperscript{1}The Hong Kong University of Science and Technology
 \textsuperscript{2}National University of Singapore\\
 \textsuperscript{3}The Hong Kong Polytechnic University\\
 \textsuperscript{4}Beijing University of Posts and Telecommunications\\
 \textsuperscript{5}Peking University
\\
 \small{
   {ycuibd@connect.ust.hk}
   {\{siruihan, yikeguo\}@ust.hk}
 }
}

\begin{document}
\maketitle

\begin{abstract}
The advent of multi-modal large language models (MLLMs) has greatly advanced research on video fake news detection (VFND) tasks. 
Existing benchmarks typically focus on the detection accuracy, while failing to provide fine-grained assessments for the entire detection process. 
To address these limitations, we introduce {POVFNDB (Process-oriented Video Fake News Detection Benchmark)}, a process-oriented benchmark comprising 10 tasks designed to systematically evaluate MLLMs' perception, understanding, and reasoning capabilities in VFND. 
This benchmark contains \textit{36,240} human-annotated question-answer (QA) in structured or open-ended formats, spanning 15 distinct evaluation dimensions that characterize different aspects of the video fake news detection process.
Using POVFNDB, we conduct comprehensive evaluations on both proprietary and open-source MLLMs. 
Moreover, we establish a strong benchmark baseline by fine-tuning Qwen2.5VL-7B-Instruct on process-oriented chain-of-thought data constructed with our proposed POVFND-CoT framework, achieving state-of-the-art performance on VFND.

\end{abstract}

\section{Introduction}
With the rise of social media, video has emerged as an important medium for news dissemination.
However, this shift facilitates the spread of fake news due to the high pervasion of video fake news \cite{nan2024let, qi2023two, d2021fake}, posing risks to social stability and necessitating effective video fake news detection (VFND).

\begin{figure}[t]  
    \hspace{-10pt}
    \includegraphics[scale=0.38]{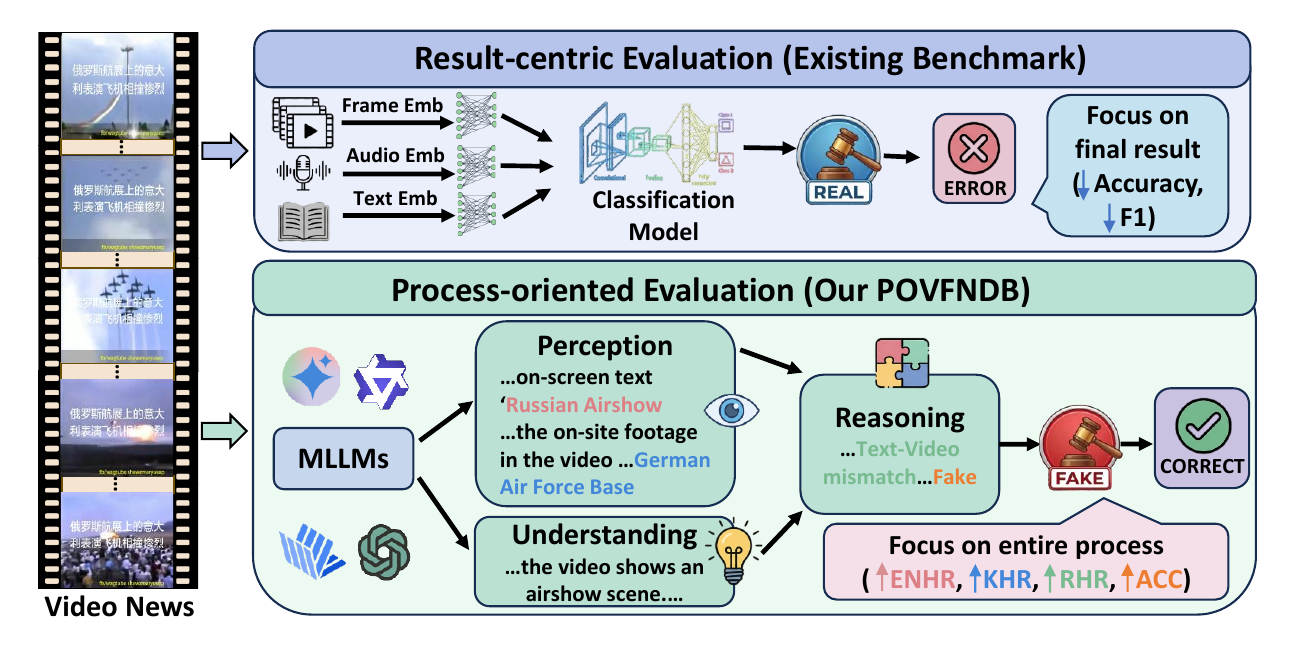}  
    \captionsetup{justification=raggedright, singlelinecheck=false}
    \caption{Comparison between result-centric and process-oriented evaluation. The MLLM correctly identifies entities (red), retrieves knowledge (blue), applies rationales (green), and reaches conclusions (orange), with all corresponding metrics (matching font colors) improving.}
    \label{intro_chart_vfnd}
    \vspace{-17pt}
\end{figure}

Most existing detection paradigms \cite{zong2024unveiling, qi2023fakesv, bu2024fakingrecipe, shen2025multi} are result-centric, focusing primarily on final detection accuracy rather than the intermediate reasoning process, as illustrated in Figure~\ref{intro_chart_vfnd}.
However, fake news detection is inherently an evidence-based reasoning task rather than a simple black-box classification problem. It requires models to perform step-by-step reasoning grounded in evidence drawn from both external knowledge sources and news content to produce interpretable predictions.
As a result, explainable fake news detection has recently emerged as an important research direction.

Compared to text-based fake news, video fake news is more challenging to model and evaluate, as videos convey information through multiple modalities with higher information density and complex temporal structures. Misleading cues can arise not only from semantic content but also from editing patterns and cross-modal interactions.
Indeed, video fake news detection is a multi-faceted task that requires diverse capabilities beyond final veracity prediction. At the {\bf perception} level, models must accurately recognize and localize key elements in videos, and distinguish between intrinsic video content and creator-added elements, such as overlaid text and shooting transitions, which often play a critical role in misleading narratives.
At a higher level of {\bf understanding} and {\bf reasoning}, VFND further requires models to comprehend the underlying news events and perform evidence-based factual reasoning by integrating external knowledge with video content.
These requirements naturally align with the strengths of MLLMs, which leverage extensive world knowledge, multimodal perception and understanding, and semantic reasoning to generate coherent explanations.
However, existing benchmarks provide only final veracity labels for classification \cite{qi2023fakesv, wang2025fmnv}. The absence of supervision signals for intermediate steps prevents fine-grained assessment of the distinct capabilities involved in the VFND process, making it difficult to systematically analyze and optimize MLLMs for this task.

\begin{figure*}[t]  
    \centering
    \includegraphics[width=\textwidth]{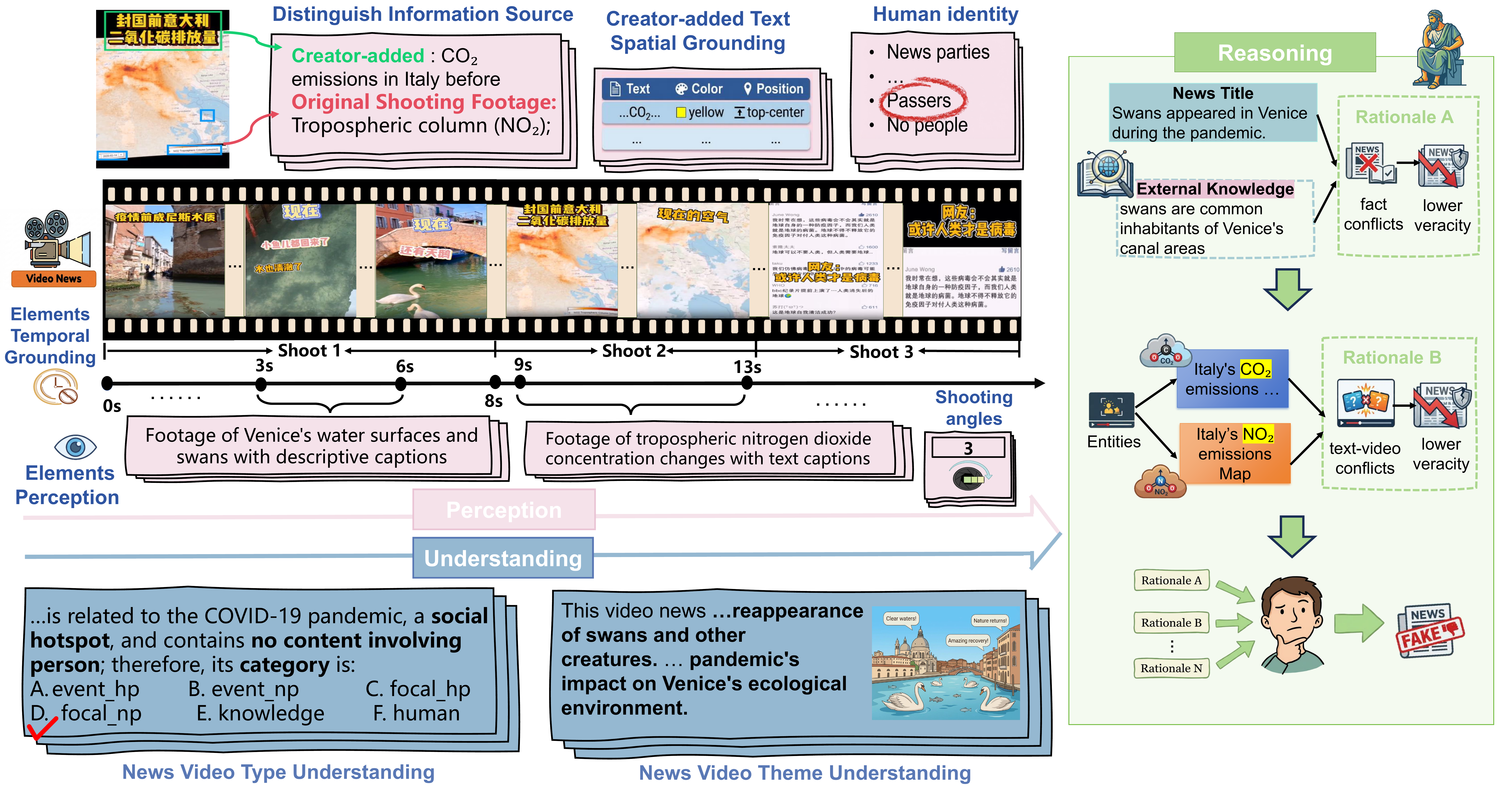}  
    \captionsetup{justification=centering, singlelinecheck=false}
    \caption{Framework of POVFNDB: Ten Evaluation Tasks Across Perception, Understanding and Reasoning.} 
    \label{framework_vfnd} 
    \vspace{-10pt}
\end{figure*}
\begin{table}[t]
\small
\setlength{\tabcolsep}{3pt}
\begin{tabular}{@{}lcccccc@{}}
\toprule
\textbf{Dataset} & \textbf{Cls} & \textbf{Per} & \textbf{Und} & \textbf{Rea} & \textbf{\#M} & \textbf{Annt (Fake/Real)} \\
\midrule
CHECKED & \textcolor{green}{\checkmark} & \textcolor{red}{$\times$} & \textcolor{red}{$\times$} & \textcolor{red}{$\times$} & 1 & 1.8k/0.34k \\
FakeSV & \textcolor{green}{\checkmark} & \textcolor{red}{$\times$} & \textcolor{red}{$\times$} & \textcolor{red}{$\times$} & 1 & 1.8k/1.8k \\
TRUE & \textcolor{green}{\checkmark} & \textcolor{red}{$\times$} & \textcolor{red}{$\times$} & \textcolor{green}{\checkmark} & 1 & 1.1k/1.8k \\
FakeVV & \textcolor{green}{\checkmark} & \textcolor{red}{$\times$} & \textcolor{red}{$\times$} & \textcolor{green}{\checkmark} & 4 & 5.1k/5.1k \\
FMNV & \textcolor{green}{\checkmark} & \textcolor{red}{$\times$} & \textcolor{red}{$\times$} & \textcolor{red}{$\times$} & 3 & 1.5k/0.9k \\
\midrule
\textbf{POVFND} & \textcolor{green}{\checkmark} & \textcolor{green}{\checkmark} & \textcolor{green}{\checkmark} & \textcolor{green}{\checkmark} & 15 & 1.8k$\times$10/1.8k$\times$10\\
\bottomrule
\end{tabular}
\captionsetup{justification=raggedright, singlelinecheck=false}
\caption{Comparison of datasets for VFND. \textbf{Cls}: Classification, 
\textbf{Per}: Perception, \textbf{Und}: Understanding, \textbf{Rea}: Reasoning,
\textbf{\#M}: Number of metrics, \textbf{Annt}: Annotations.}
\label{tab:dataset_comparison}
\vspace{-20pt}
\end{table}

To address these challenges, we develop POVFNDB, a process-oriented benchmark that supports fine-grained assessment of the diverse capabilities involved in video fake news detection.
Built upon 3,624 videos from FakeSV \cite{qi2023fakesv}, a widely used real-world video fake news dataset, POVFNDB constructs approximately 36,240 question–answer pairs organized into three core dimensions of perception, understanding, and reasoning, covering 10 sub-tasks and 15 evaluation metrics.
Based on POVFNDB, we perform a thorough evaluation of mainstream MLLMs, including open-source and proprietary video or image MLLMs. Extensive experiments demonstrate that even the leading MLLM ({\it i.e.}, Gemini2.5-Flash) exhibit subpar performance in most perception and reasoning sub-tasks, underscoring the challenges and necessity of multimodal research tailored for the VFND process.
Motivated by insights from rationale validation, we fine-tune Qwen2.5VL-7B-Instruct on a reasoning dataset generated by our proposed POVFND-CoT, a chain-of-thought method. This approach achieves state-of-the-art performance, highlighting the importance of developing domain-specific reasoning strategies for VFND. 
Our contributions can be summarized as follows: 
\begin{itemize}[itemsep=0pt, parsep=0pt, topsep=0pt]
\item \textbf{Novel Process-Oriented Benchmark}: 
The proposed POVFNDB benchmark
is organized around three core dimensions, perception, understanding, and reasoning, and comprises {10 sub-tasks} with {15 evaluation metrics} based on {36,240 question–answer annotations}. 
This design enables fine-grained evaluation and diagnostic analysis of model capabilities throughout the video fake news detection pipeline.
\item \textbf{Comprehensive Evaluation of MLLMs}: 
Using POVFNDB, we extensively evaluate mainstream MLLMs, including open-source/proprietary and image-/video-based models, revealing their strengths and limitations across VFND sub-tasks and diverse capability dimensions.
\item \textbf{Strong Explainable Detection Baseline}: 
We further propose {POVFND-CoT}, an explainable chain-of-thought framework for VFND that leverages process-aware reasoning signals. 
By fine-tuning models on reasoning data generated under POVFND-CoT, we establish a strong baseline that achieves state-of-the-art performance on video fake news detection.
\end{itemize}

\section{Related Work}
\subsection{Video Fake News Detection}
Fake News Detection (FND) seeks to evaluate the veracity of news by analyzing characteristics across multiple modalities \cite{xu2025triplefact, zhang2025knowledge, guo2025each, wang2024explainable}, including text, images, audio, and video. \cite{zhou2019network}leveraged network-based features and propagation patterns to enhance fake news detection for textual articles. Sniffer\cite{qi2024sniffer} leverages MLLMs to provide explainable misinformation detection for image-based news.
Video news convey richer information\cite{wang2025fakesv, zong2025text, zeng2025imol, wu2023mfir}.\cite{qi2023fakesv} demonstrated that integrating video content, comments, and metadata collected from social media platforms significantly enhance the effectiveness of VFND.

\subsection{Video Benchmark for MLLM Evaluation}
The emergence of extensive multi-modal datasets \cite{yuan2025survey} and architectural innovations have significantly propelled the development of MLLMs. These advancements have enabled MLLMs to achieve competitive performance in core multi-modal tasks \cite{tang2025video, wang2025multimodal}. Videos exhibit inherent temporal dynamics, such as motion patterns \cite{qian2024streaming, wang2024videoagent} and color transitions, offering unique analytical features that distinguish them from static text and image modalities. There has been growing interest in exploring MLLMs' capacities for video-based tasks \cite{ qian2024momentor, tang2025cardiff, feng2025video}. Video-of-Thought \cite{fei2024video} introduced spatial-temporal grounding-aware tuning, which effectively bridges perceptual insights with cognitive reasoning. 
As MLLMs continue to demonstrate robust multi-modal video processing performance, a variety of benchmarking \cite{ning2023video, zhang2025q, caba2015activitynet, liu2024bench} have been developed to evaluate their perception, understanding, and reasoning capacities. MMMU \cite{yue2024mmmu} evaluates visual perception and reasoning abilities across a broad range of disciplines. MMBench-Video \cite{fang2024mmbench} incorporates long-form videos to assess MLLMs' spatial-temporal understanding. 

However, these benchmarks focus on general-purpose multi-modal capabilities \cite{han2025video, sun2025ve, wang2024sok}, overlooking specialized domain tasks. 
As a comparison, our proposed benchmark is designed to specifically evaluate MLLMs' discriminative visual perception, content understanding, and veracity reasoning abilities in the video fake news detection task.

\section{Proposed Benchemark}
\begin{table*}[t]
    \centering
    \small
    \begin{tabular}{@{}clp{3.5cm}@{\hspace{0.3cm}}ccc@{}}  
        \toprule
        \textbf{Task Type} & \textbf{Task} & \textbf{Format} & \textbf{Target} & \textbf{Metrics} \\
        \midrule
        \multirow{7}{*}{\centering Perception} & Key Elements Perception (KEP) & Open-Ended with format & OSF & Avg.EHR \\
        & Distinguish CAC and OSF (DCS) & Open-Ended with format & CAC\&OSF & ROUGE-L \\
        & Creator-added Text Color (CCP) & Single Selection & CAC & Accuracy \\
        & Creator-added Text Position (CPP) & Single Selection & CAC & Accuracy \\
        & Key Elements Grounding (KEG) & Open-Ended with format & OSF & IoU \\
        & Shooting Angles Counting (SAC) & Open-Ended with format & OSF & Avg.AD \\
        & Human Identity Recognition (HIR) & Multi Selection & OSF & Accuracy \\
        \midrule
        \multirow{2}{*}{\centering Understanding} & News Video Type (NTU) & Open-Ended & CAC\&OSF & Accuracy. \\
        & News Video Theme (NEU) & Single Selection & CAC\&OSF & Avg.FC/TR/CO \\
        \midrule
        \multirow{4}{*}{\centering Reasoning} & \multirow{4}{*}{Final Detection Reasoning (FDR)} & \multirow{4}{*}{Open-Ended} & CAC\&OSF & Avg.ENHR \\
        & & & Ext.Know. & Avg.KHR \\
        & & & Rationale & Avg.RHR \\
        & & & Real/Fake & Accuracy \\
        \bottomrule
    \end{tabular}
    \caption{Task Description. {CAC}: Creator-added content. {OSF}: Original shooting footage. {Ext.Know.}: External Knowledge. {Avg.EHR}: Average elements hit rate. {Avg.AD}: Average absolute distance. {Avg.ENHR}: Average entities hit rate. {Avg.RHR}: Average rationale hit rate. {Avg.KHR}: Average knowledge hit rate. {FC}: Factual consistency. {TR}: Theme relevance. {CO}: Completeness.}
    \label{details4task_and_data}
    \vspace{-12pt}
\end{table*}
In this section, we detail the construction of POVFNDB. We first introduce the definitions and motivations of the process tasks, then describe video selection and annotation procedures, present the resulting data distributions, and finally outline the task-specific evaluation methodology.
\subsection{Task definition}
Following practical detection workflows, we design tasks targeting different stages of the verification process. Perception tasks target the stage of visual evidence extraction from both creator-added content(CAC) and original shooting footage(OSF). Understanding tasks target the global video news comprehension. Reasoning tasks target the stage where MLLMs synthesize evidence to produce verification judgments. In this section, we present formal task definitions and discuss their necessity in the VFND. All tasks are annotated by the process in Figure \ref{data_construct_vfnd}, and concrete examples of these tasks are provided in Appendix \ref{label_task_definitions}.
\begin{figure}[t]  
    \centering
    \includegraphics[scale=0.53]{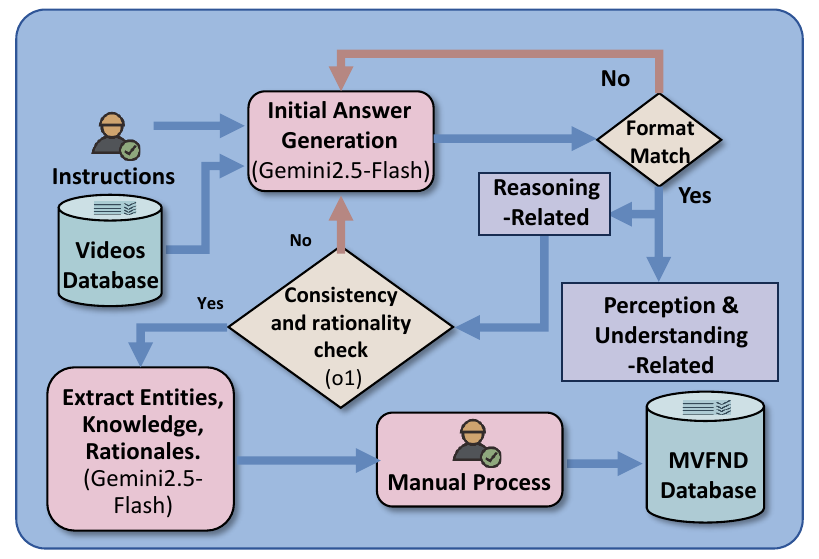}  
    \captionsetup{justification=centering}
    \caption{POVFND Dataset Construction.}
    \label{data_construct_vfnd}
    \vspace{-12pt}
\end{figure}
\subsubsection{Perception}
\begin{figure}[t]  
    \includegraphics[scale=0.45]{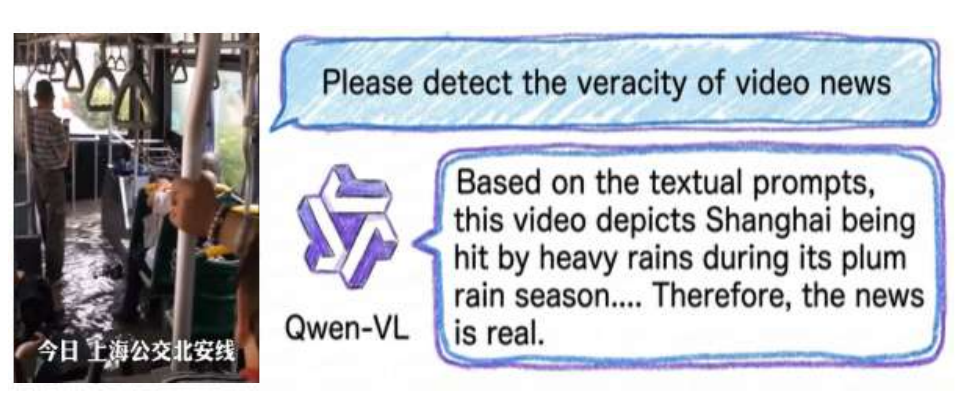}  
    \captionsetup{justification=raggedright, singlelinecheck=false}
    \caption{When identifying video veracity, MLLMs uncritically accept the text added by creators, which leads to erroneous conclusions.}
    \label{dis_cac_osf}
    \vspace{-10pt}
\end{figure}

\begin{figure}[t]  
    \hspace{-8pt}
    \includegraphics[scale=0.35]{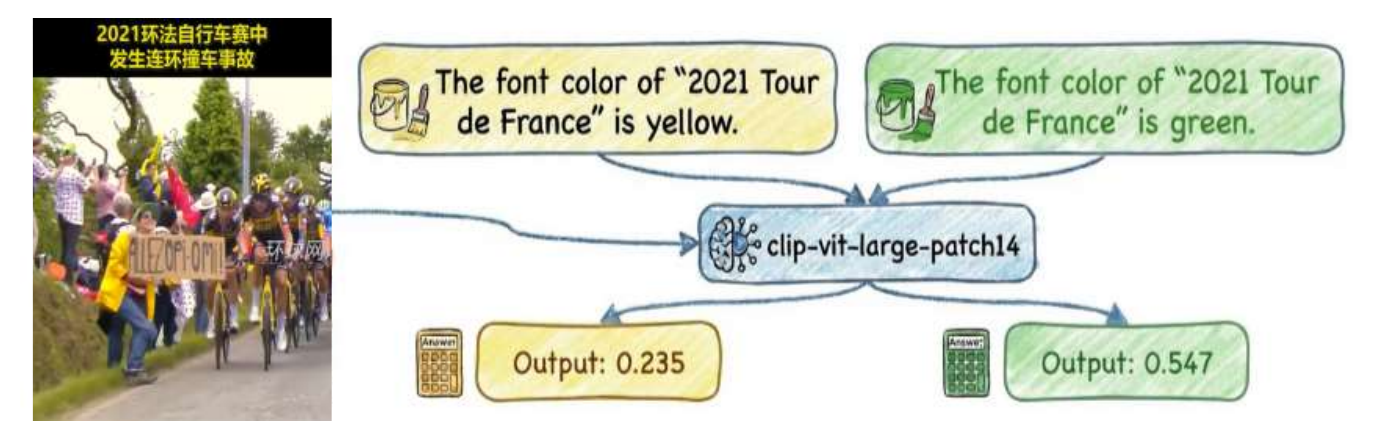}  
    \captionsetup{justification=raggedright, singlelinecheck=false}
    \caption{CLIP model misclassifies text color in video; green is given higher similarity.}
    \label{clip_cat_color}
    \vspace{-15pt}
\end{figure}

\textbf{Key Elements Perception (KEP).} The task requires MLLMs to identify all elements in original shooting footage that are crucial for VFND. These elements include on-site footage, interview segments, and other components. These elements offer primary source information about the events.\\
\textbf{Distinguish Creator-added Content and Original Shooting Footage (DCS).} 
The task involves recognizing creator-added content and original shooting footage in video news. The former is more susceptible to manipulation and carries stronger authorial intent, whereas the latter provides relatively unaltered information. However, MLLMs usually treat both types as equally credible, as they are pretrained on semantic features rather than provenance distinctions, as illustrated in Figure \ref{dis_cac_osf}. \\
\textbf{Creator-added Text Color Perception (CCP).} In this task, MLLMs are required to perceive the font color of text in videos. Text color constitutes a crucial component of news, as it reflects the creator's professional background and influences comprehension \cite{zhou2022effects}. However, previous experiments reveal that Clip-based MLLMs struggle to identify color of visual content, as the error illustrated in Figure \ref{clip_cat_color}.\\
\textbf{Creator-added Text 2D-position Perception (CPP).} Similar to text color, text position reflects the creation style of news content, yet MLLMs face limitations in recognizing spatial locations of text. \\
\textbf{Shooting Angles Counting(SAC).} More camera angles enable multi-perspective record of news events, particularly in accident scene footage. Thus, camera angles constitute a crucial attribute of OSF. This task requires MLLMs to identify angle transition and count the number of distinct viewpoints.\\
\textbf{Human Identity Recognition (HIR).} Given a video, MLLMs were required to retrieve and identify all identities that have appeared in the video.\\
\textbf{Key Elements Temporal Grounding (KEG). } By leveraging the temporal ranges of key elements, MLLMs conduct more precise visual information extraction and verification. Consequently, KEG enhances MLLMs' video understanding accuracy, thereby improving overall VFND performance.
\subsubsection{Understanding}
\textbf{News Video Type Understanding (NTU).} Different news types rely on distinct evidence for verification, as illustrated in the Figure \ref{combined_radar_reason_entity_knowledge_rationale}. Therefore, determining video type before applying type-specific detection strategies can effectively improve both VFND efficiency and accuracy.\\
\textbf{News Video Theme Understanding (NEU).} This task requires MLLMs to obtain a global comprehension of video news and extract thematic information (What, When, Where, Who, How, etc.). This information guides MLLMs to focus on elements relevant to the key content.
\subsubsection{Reasoning}
\textbf{Final Detection Reasoning (FDR).} This task requires MLLMs to synthesize multi-modal entities extracted through perception and understanding capabilities, incorporate general knowledge, and apply logical rationales to reason video veracity.

To comprehensively evaluate MLLMs' capabilities, we employ task-specific evaluation strategies tailored to output formats, as shown in Table \ref{details4task_and_data}. For structured outputs, we use rule-based exact matching metrics(e.g.,IoU). For instance, in the CCP task, we assess performance via perception accuracy. For tasks requiring holistic comprehension of global video semantics, such as news theme understanding, we adopt semantic-based metrics including factual consistency, relevance, and completeness, evaluated through GPT-4o as an automated judge.

\subsection{Dataset Construction and Statistics}

\textbf{Data Collection.} 
We collect videos from FakeSV, which contains real-world news videos from popular social media platforms across diverse domains. This dataset offers two key advantages: (1) labels are cross-validated against official fact-checking platforms, ensuring reliability; (2) videos retain authentic social media characteristics, enabling our benchmark to capture realistic fake news patterns and align with practical VFND. To ensure unbiased evaluation, we construct a balanced test set with 753 video pairs across 6 distinct news categories, knowledge-based news account for 13.4\%.

\noindent\textbf{Data Annotation.} 
For data annotation, as depicted in the Figure \ref{data_construct_vfnd}, to reduce bias and hallucination, we utilize Gemini2.5-flash with strong video abilities to generate and refine visual tasks. Then, we use O1 to evaluate and filter the reasoning outputs of MLLMs. The O1 model evaluates these reasoning outputs for factual consistency and logical rationality, assigning scores from 0 to 5. Outputs scoring below 4 in either metric are reprocessed by the Gemini. Finally, perception generations are verified by 20 news creation experts after 10 hours annotation training; understanding and reasoning outputs are reviewed and revised by 10 detection experts (with 5+ years of video news censor experience in online public opinion regulatory authority) and 5 domain experts(with 5+ years of specialized field  research experience). This iterative process continues until all task data is processed.

\begin{figure}[t]  
    \hspace{-10pt}
    \includegraphics[scale=0.53]{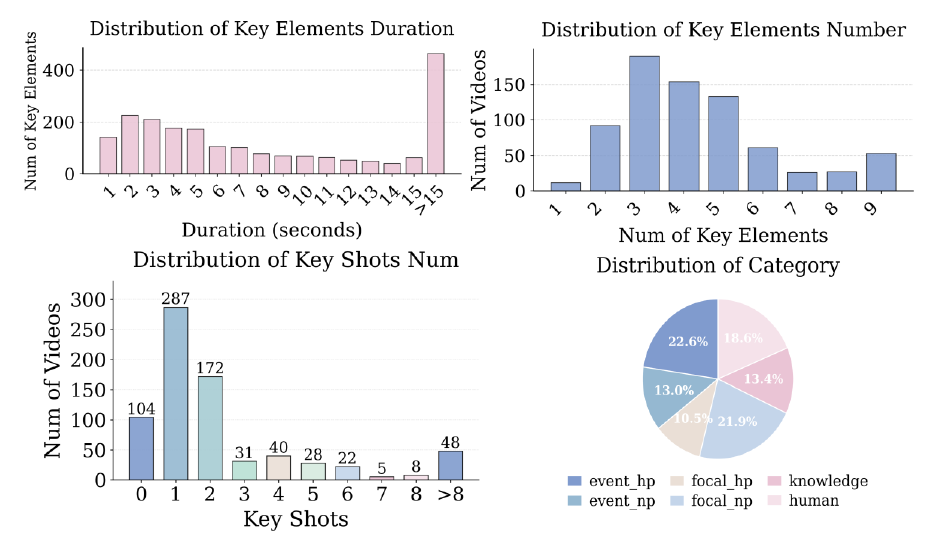}  
    \captionsetup{justification=centering}
    \caption{POVFND Dataset Statistics.}
    \label{data_statistics_test_dataset}
    \vspace{-14pt}
\end{figure}

\noindent\textbf{Data Statistics.}
POVFNDB comprises 3,624 videos across 6 news categories, split into 753 test videos and 2,871 training videos. Each video is accompanied by visual information, semantic descriptions, and reasoning evidence. As shown in Figure \ref{data_statistics_test_dataset}, the test split contains an average of 4.40 key elements per video, totaling 2,875 key elements, such as on-site footage. The duration of key elements and the number of key shots exhibit diverse distributions. The dataset maintains balanced representation across 6 news categories.

\subsection{Evaluation Metrics}
\label{lab_metrics}
To evaluate MLLM performance across the 10 tasks, we define task-specific metrics tailored to output formats and evaluation objectives. In this section, we present the mapping between metrics and tasks, with detailed calculation procedures provided in Appendix \ref{appendix_lab_metrics}. \\
\textbf{Avg. EHR (Element Hit Rate):} Measures the element hit rate in the KEP task, assessing MLLMs' ability to identify critical visual information in news videos (e.g., on-scene footage).\\
\textbf{Avg. ENHR (Entity Hit Rate):} Measures the entity hit rate in the FDR task, assessing MLLMs' ability to cite video entities during reasoning (e.g., on-screen text for locations and identities).\\
\textbf{Avg. KHR (Knowledge Hit Rate):} Measures the knowledge hit rate in the FDR task, assessing MLLMs' ability to leverage external knowledge during reasoning (e.g., physics, economics).\\
\textbf{Avg. RHR (Rationale Hit Rate):} This metric measures the rationale hit rate in the FDR, evaluating MLLMs' ability to apply reasoning logic during the process, such as contradictions between video content and facts that reduce credibility.\\
\textbf{Avg. AD (Absolute Distance):} Measures the absolute distance between predicted and ground truth shooting angle counts in the SAC task. Lower values indicate better performance.\\
\textbf{Avg. IoU (Intersection over Union):} Measures temporal grounding accuracy in the KEG by evaluating MLLMs' ability to localize multiple key elements per video. Higher values indicate better localization performance.

\section{Experiments}
\begin{table*}[!t]  
    \centering
    \renewcommand{\arraystretch}{1.3}  
    \resizebox{\linewidth}{!}{
        \begin{tabular}{l|ccccccc!{\vrule width 0.8pt}cccc}
            \hline
            \multirow{2}{*}{\diagbox[width=12em,height=3.2em]{\hspace{2em}\textbf{Model}}{\vspace{-3em}\hspace{-3em}\textbf{Task}}} & \multicolumn{7}{c!{\vrule width 0.8pt}}{\textbf{Perception}} & \multicolumn{4}{c}{\textbf{Understanding}} \\
            \cline{2-12}
            & KEP & DCS & CCP & CPP & KEG & SAC & HIR & NTU & NEU(FC) & NEU(TR) & NEU(CO) \\
            \hline
            Gemini2.5-Flash & \best{66.95} & \best{67.34} & \best{47.47} & \best{59.80} & \best{0.3819} & \best{1.88} & \best{78.72} & \best{49.72} & \best{4.36} & \best{4.40} & \best{3.8} \\
            GPT-4o-mini      & 55.59 & 34.78 & \secondbest{45.26} & 28.51 & 0.1573 & 4.22 & 58.93 & 42.37 & 3.59 & 3.60 & 2.13 \\
            Qwen2.5-VL-72B-Instruct    & \secondbest{59.61} & \secondbest{60.26} & 39.15 & \secondbest{50.36} & \secondbest{0.2346} & \secondbest{2.03} & \secondbest{77.27} & \secondbest{44.86} & \secondbest{4.04} & \secondbest{4.08} & \secondbest{2.63} \\
            Qwen2.5-VL-32B-Instruct    & 54.30 & 57.03 & 31.71 & 35.73 & 0.2218 & 2.16 & 70.47 & 38.67 & 3.62 & 3.66 & 2.27 \\
            Qwen2.5-VL-7B-Instruct     & 52.86 & 40.86 & 28.33 & 24.82 & 0.1589 & 2.71 & 19.38 & 30.84 & 3.37 & 3.51 & 1.87 \\
            InternVL3-78B    & 57.39 & 58.65 & 44.61 & 40.48 & 0.1553 & 2.24 & 59.22 & 41.12 & 3.58 & 3.63 & 2.30 \\
            InternVL3-38B    & 53.26 & 58.37 & 30.36 & 31.28 & 0.1663 & 2.28 & 61.39 & 36.95 & 3.45 & 3.41 & 2.08 \\
            InternVL3-8B    & 51.35 & 38.17 & 25.36 & 45.74 & 0.1312 & 2.86 & 16.52 & 28.15 & 3.28 & 3.32 & 1.62 \\
            \hline
        \end{tabular}
    }
    \caption{Evaluation on Perception and Understanding. Best results are in \textbf{bold} with darker background, second best with lighter background.}
    \vspace{-12pt}
    \label{perception_understanding_tab}
\end{table*}

\begin{table}[!t]
    \centering  
    \renewcommand{\arraystretch}{1.4}  
    \resizebox{\linewidth}{!}{
        \begin{tabular}{@{}l!{\vrule width 0.8pt}cccc@{}}
            \hline
            \multirow{2}{*}{\diagbox[width=14.5em,height=3.4em,trim=l,innerrightsep=0pt,innerleftsep=0pt]{\hspace{2em}\textbf{Model}}{\hspace{-5em}\textbf{Task}}} & \multicolumn{4}{c}{\textbf{Reasoning}} \\
            \cline{2-5}
            & \textbf{ENHR} & \textbf{RHR} & \textbf{KHR} & \textbf{ACC} \\
            \hline
            Gemini2.5-Flash & \best{62.67} & \best{79.57} & \best{47.11} & \best{77.29} \\
            GPT-4o-mini     & 31.74 & 41.26 & 24.75 & 69.72 \\
            Qwen2.5-VL-72B-Instruct  & \secondbest{46.62} & \secondbest{68.76} & \secondbest{37.59} & \secondbest{75.70} \\
            Qwen2.5-VL-32B-Instruct  & 30.14 & 61.36 & 29.85 & 73.17 \\
            Qwen2.5-VL-7B-Instruct   & 25.35 & 40.47 & 22.73 & 68.79 \\
            InternVL3-78B   & 25.32 & 33.09 & 23.61 & 74.50 \\
            InternVL3-38B   & 21.17 & 30.05 & 21.51 & 71.71 \\
            InternVL3-8B    & 18.73 & 26.73 & 18.64 & 68.26 \\
            \hline
        \end{tabular}
    }
    \caption{Evaluation on Final Detection Reasoning.}
    \vspace{-12pt}
    \label{label_final_reason_result}
\end{table}
\begin{table}[t] 
    \centering 
    \renewcommand{\arraystretch}{1.4}
    \resizebox{1.05\linewidth}{!}{
        \normalsize
        \begin{tabular}{@{}>{\centering\arraybackslash}p{2.5em}|@{\hspace{0.5em}}>{\raggedright\arraybackslash}p{7em}!{\vrule width 0.8pt}lccccc@{}}
            \hline
            \multicolumn{2}{l!{\vrule width 0.8pt}}{\multirow{2}{*}{%
            \begin{tikzpicture}[baseline=(current bounding box.center)]
                \node[minimum width=8em, minimum height=2.4em, inner sep=0pt] (box) {};
                \draw[shorten >=-28pt, shorten <=-8pt] ([yshift=8pt]box.north west) -- ([yshift=8pt]box.south east);
                \node[anchor=south west, inner sep=2pt] at (box.south west) {\textbf{Method}};
                \node[anchor=north east, inner sep=2pt] at (box.north east) {\textbf{Task}};
            \end{tikzpicture}%
            }} & \multicolumn{5}{c}{\textbf{Qwen2.5VL-7b-Instruct}}\\
            \cline{3-7}
            \multicolumn{2}{l!{\vrule width 0.8pt}}{} & \textbf{ACC} & \textbf{M-F1} & \textbf{ENHR} & \textbf{RHR} & \textbf{KHR}\\ 
            \hline
            \multicolumn{2}{l!{\vrule width 0.8pt}}{Zero-Shot} & 68.79 & 67.41 & 25.35 & 40.47 & 22.73 \\
            \hline
            \multirow{4}{=}{CoT} & POVFND & 70.14 & 68.54 & 28.37 & 41.57 & 24.39 \\
            & CAC-POVFND & 71.58 & 70.64 & 32.15 & 45.65 & 26.71 \\
            & OSF-POVFND & \secondbest{73.87} & \secondbest{72.38} & 33.17 & \secondbest{49.26} & \secondbest{29.71} \\
            & ALL-POVFND & 72.64 & 71.59 & \secondbest{40.13} & 48.31 & 28.10 \\
            \hline
            \multicolumn{2}{l!{\vrule width 0.8pt}}{Instruct-Tuning} & \best{81.14} & \best{80.26} & \best{74.58} & \best{83.15} & \best{51.37} \\
            \hline
        \end{tabular}
    }
    \caption{Evaluation results of different CoT variants on VFND using Qwen2.5VL-7b-Instruct.}
    \label{cot_reason_result}
    \vspace{-12pt}
\end{table}
Utilizing POVFNDB, we evaluate a diverse set of MLLMs, including both video-based and image-based models from open-source and proprietary source. 
Overall evaluation results are presented in Table \ref{perception_understanding_tab} and \ref{label_final_reason_result}. Beyond the process-oriented tasks evaluation, we select rationales in MLLMs' reasoning outputs, conducting in-depth analysis within video visual features to verify validity of these rationales.
Based on these findings, we propose POVFND-CoT (Table \ref{prompt4_mvfnd_cot}) to assess performance gains from feature integration on detection accuracy. We further fine-tune Qwen2.5VL-7B-Instruct on the reasoning outputs generated by POVFND-CoT. The fine-tuned model achieves sota performance across 4 reasoning evaluation metrics.  
\subsection{Experimental Settings}
We conduct comprehensive experiments on 10 tasks with 8 MLLMs in our benchmark, encompassing  both open-source and proprietary models. For image MLLMs, we leverage their inherent image process capacities to utilize evenly sampled 16 frames from each news video. For video MLLMs, we utilize a uniform sampling method or models' predefined sampling strategy for video processing. We employ models' official implementations or available APIs for evaluation. We assess both backbone and fine-tuned models with zero-shot manner.

\subsection{Results Analysis}
\begin{table*}[!t]
\centering
\small
\setlength{\tabcolsep}{4pt}
\begin{tabular}{>{\raggedright\arraybackslash}m{2.8cm}m{12.7cm}}
\toprule
\textbf{News Video} & 
\includegraphics[width=12.5cm,height=3cm]{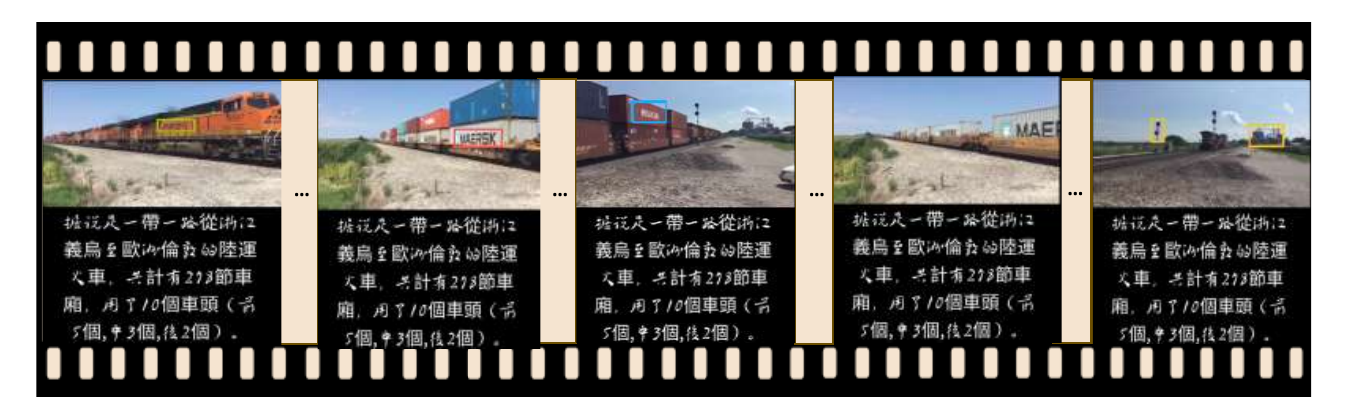} \\
\midrule
\textbf{Title} & 
Belt and Road Land Freight Train from Yiwu, Zhejiang to London, UK \\
\midrule
\textbf{Original Footage \& Creator Content} & 
\makecell[l]{\textbf{Original footage:} Brands on the train. ...train moving through rural landscape.... \\
\textbf{Creator-added content:} ``\textcolor{red}{Belt and Road land transport train from Yiwu to London}''. }\\
\midrule
\textbf{Creator-added Text} & {\bf Content:} Belt and Road ... from Yiwu... to London,...; {\bf Color:} white; {\bf Position}: center and bottom center.\\
\midrule
\textbf{Key Elements Perception} & 
Branding on train engines: BNSF, HYUNDAI and MAERSK logos. \textbf{Environmental context:} Landscape, vegetation, railway infrastructure... \\
\midrule
\textbf{Key Elements Grounding \& Analysis} & 
\textbf{20s-30s (BNSF):} ... \textcolor{red}{``BNSF'' logo}. BNSF = Burlington Northern Santa Fe, a \textcolor{red}{major North American railroad}... \textcolor{red}{does not operate on Eurasian land bridge}... \textbf{General landscape:} ... railway infrastructure aligns with \textcolor{red}{North American standards}... \\
\midrule
\textbf{Identity Recognition} & 
No specific individuals (interviewees, officials, reporters) appear. Purely train and landscape footage. \\
\midrule
\textbf{Shooting Angles} & 
Two angles capturing train passing by, likely from fixed position or drone. \\
\midrule
\textbf{POVFND-CoT Veracity Reasoning} & 
\textbf <think>...the "BNSF" logo belongs to Burlington Northern Santa Fe, \textcolor{red}{a major North American railroad operating exclusively in the United States and Canada, not on China-Europe routes ... HYUNDAI and MAERSK markings, neither of which operates direct rail services from Yiwu to London}. ..., \textcolor{red}{railway signals, and silos characteristic of the North American Midwest, not the Eurasian...}... The text's use of "supposedly" is highly unusual for factual news reporting, indicating unverified sources. \textcolor{red}{The non-standard font and intrusive text layout occupying half the screen further deviate from conventional news presentation standards.} The fundamental discrepancy between the BNSF North American footage and the claimed China-Europe narrative, combined with stylistic red flags, clearly identifies this as fake news through content misappropriation</think><result>\textcolor{blue}{Fake}</result> \\
\midrule
\textbf{Ground Truth} & Fake \\
\bottomrule
\end{tabular}
\caption{Case study demonstrating POVFND-CoT's step-by-step reasoning process.}
\label{cot_case_study}
\vspace{-12pt}
\end{table*}
\textbf{MLLMs exhibit poor performance on creator-added text tasks (CCP and CPP),} with the best model achieving below 60\% accuracy on both tasks. We attribute this to MLLM training paradigms that prioritize semantic understanding while neglecting spatial positioning and visual attributes (e.g., font color). This reveals a critical gap in existing MLLMs for news video perception, where text-rich visual content is prevalent. Our results suggest that specialized training targeting text perception abilities is essential for effective MLLM-based VFND.\\
\textbf{Temporal grounding across multiple key elements challenges model detection performance.} Only Gemini-2.5-Flash exceeded the standard threshold of 0.3, indicating that dense temporal grounding in news videos remains challenging for off-the-shelf MLLMs. Figure \ref{model_key_elements_num_accuracy_chart} demonstrates that all models achieve reasonable performance with a single element. However, as the number of key elements increases, only models (e.g., Gemini2.5-Flash) with stronger localization abilities show performance gains. Notably, we observe a positive correlation between multi-element grounding capability and VFND accuracy.\\
\textbf{Video-based MLLMs demonstrate better dynamic feature capturing capabilities.} For instance, Qwen2.5-VL-7B achieves comparable temporal grounding performance to GPT-4o-mini while exhibiting stronger SAC capabilities, despite its smaller size. We attribute the advantage to Qwen2.5-VL's native video processing architecture and dynamic FPS sampling mechanism, which enable more effective temporal modeling. These findings suggest that image-based processing paradigms limit MLLMs' visual abilities and VFND performance.\\
\textbf{Hit rate metrics provide evidence-based reasoning verification.}
As demonstrated in Figure \ref{label_final_reason_result}, hit rates exhibit strong positive correlation with detection accuracy, with improvements in hit rates consistently accompanied by accuracy gains.
These patterns align with real-world verification practices, indicating that our benchmark closely reflects practical detection workflows.\\
\textbf{OSF features provide larger improvements than CAC.} As shown in Table \ref{cot_reason_result}, we attribute this to two factors. First, OSF captures dynamic visual elements with richer contextual information, while CAC is limited to semantic with a constrained feature space. Second, OSF originates from authentic footage and is less susceptible to manipulation, whereas CAC represents author-added content with lower credibility. Consequently, models rely more heavily on OSF, yielding superior detection performance. This suggests that MLLM-based VFND should prioritize exploring discriminative patterns in OSF when conduct authentic VFND.\\
\subsection{Further Analysis}
Based on evaluation results and rationale validation, we propose POVFND-CoT (Table \ref{prompt4_mvfnd_cot}), which tailors reasoning paths based on news type: knowledge-oriented, content-oriented, or hybrid. This CoT approach enables MLLMs to cite external knowledge, perceive target visual features, and apply detection rationales to assess veracity. To examine how different video features contribute to detection performance, we develop CoT variants combining POVFND with CAC and OSF:
\textbf{CAC-POVFND-CoT :} POVFND integrates creator-added content analysis (color, position, text), details as shown in Appendix Table \ref{prompt4_cac_mvfnd_cot}.
\textbf{OSF-POVFND-CoT :} POVFND incorporates original shooting footage analysis (key elements, shooting angles, human identities), details as shown in Appendix Table \ref{prompt4_osf_mvfnd_cot}.
\textbf{ALL-POVFND-CoT :} POVFND integrates both CAC and OSF components, details as shown in Appendix Table \ref{prompt4_all_mvfnd_cot}.

As shown in Table \ref{cot_reason_result}, we find \textbf{MLLMs struggle to integrate OSF and CAC.}, ALL-POVFND-CoT underperforms OSF-POVFND-CoT despite incorporating CAC information. We attribute this to MLLMs' inability to properly weight each feature's contribution, reflecting VFND's domain-specific reasoning abilities. 

To address this, we use ALL-POVFND-CoT to guide Gemini2.5-Flash in generating reasoning outputs. This enables MLLMs to utilize extensive visual information and perform end-to-end detection, incorporating retrieval triggers and evidence-based rationales. The outputs reveal how MLLMs weigh various evidence to derive conclusions. We fine-tune Qwen2.5-VL-7B-Instruct on these training samples (parameters in Table \ref{training_params}), with the fine-tuned model achieving state-of-the-art performance in test-split. This demonstrates that effective VFND requires specialized reasoning strategies developed through domain-specific fine-tuning rather than general capabilities alone.

\subsection{Case Study}
Does the MLLM distinguish between creator-added text and text in the original footage ? As shown in Table \ref{cot_case_study}, the model identifies text on the train cars, such as 'BNSF', as well as the creator-added text 'Belt and Road land transport train from Yiwu to London'. The MLLM utilizes these two distinct text types for different analytical purposes. For creator-added text, it examines the location, font color, and semantic content. For text appearing on the train, the model uses it to retrieve information for fact-checking.

Does the MLLM effectively leverage external knowledge to assist VFND ? The model utilizes the captured logo 'BNSF' on the train to retrieve the fact that 'BNSF is a major North American railroad operating exclusively in the United States and Canada, not on China-Europe routes'. Additionally, the MLLM analyzes the environmental context in the footage and identifies characteristics typical of the North American Midwest rather than Eurasia. These factual discrepancies with the news content significantly undermine the credibility of the news.

Does the MLLM correctly identify the news type as knowledge and content oriented ? The model's output incorporates both external knowledge verification and video content analysis, indicating that the MLLM recognizes the news type during reasoning and adapts its subsequent reasoning steps.
\section{Conclusion}
In this work, we introduce POVFNDB, a process-oriented benchmark for evaluating MLLM capabilities in VFND through comprehensive video feature descriptions and process-centric tasks. Our extensive evaluation reveals significant limitations in existing MLLMs' perception, understanding, and reasoning abilities for fake news detection. We identify key strengths and weaknesses of current MLLMs and establish that effective VFND necessitates domain-specific reasoning strategies rather than relying solely on general-purpose capabilities.
\section*{Limitations}
Our work has explored the utilization of MLLMs for VFND, however, there are also some limitations Firstly, Fake news instances of the same type may exhibit similar deceptive characteristics, as the nature of the facts they aim to obscure is comparable. This aspect, however, remains underexplored in our current work. Secondly, the application of external knowledge requires careful verification of its authenticity, which typically necessitates cross-validation from multiple sources. Further exploration of these approaches remains necessary.

\clearpage
\bibliography{custom}
\clearpage
\appendix

\section{Data Curation}
\label{lab_data_curation}
\subsection{Details of Videos Category}
\textbf{event\_hp: }This category of news primarily reports on specific incidents and their associated individuals. Typical cases include fatalities caused by electric shock, major traffic accidents, and other similar events, all of which are attributed to human factors.\\
\textbf{event\_np: }This category of news primarily focuses on reporting specific incidents without involving information about associated individuals. Typical examples include sudden natural disasters such as typhoons, earthquakes, and flash floods. These incidents have no clear direct correlation with human activities, and the core content of the reports is centered on the description and introduction of the incidents themselves.\\
\textbf{focal\_hp: }This category of news primarily focuses on reporting events related to social hotspots, with the main content centered on human subjects. A typical example is the withdrawal of a star player from a match due to injury during the FIFA World Cup—such events are closely associated with the topics under discussion in social media public opinion at the corresponding time.\\
\textbf{focal\_np: }This category of news primarily reports on events associated with social hotspots, and the content of such reports does not involve specific individuals or groups. A typical instance is the sell-out of goods in a certain supermarket prior to the arrival of a typhoon.\\
\textbf{human: }This category of news primarily focuses on profiling an individual, with its content centered on introducing a specific aspect of the individual rather than targeting a single event associated with them. A typical example is the summary of a worker's work experience. Generally, this type of news does not prioritize timeliness, as it is more inclined to present a summary of the individual's past experiences.\\
\textbf{knowledge: }This category of news primarily focuses on disseminating common knowledge, with its core content dedicated to the introduction and elaboration of such knowledge. Typical examples include the popularization of scientific knowledge, health maintenance guidance, and daily life tips. Generally, this type of news features a relatively high knowledge density, and its main body relies on verbal narration for content delivery rather than video demonstrations.\\
\begin{figure*}[!htbp]  
    \centering
    \includegraphics[width=\textwidth]{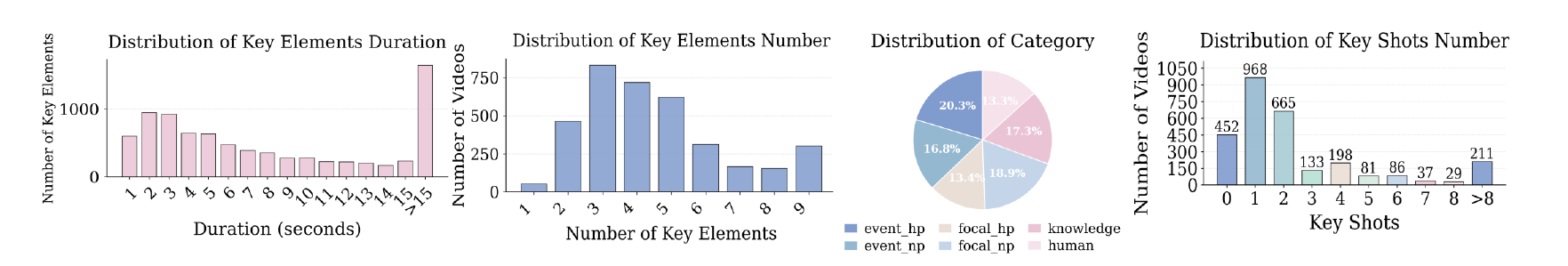}  
    \captionsetup{justification=centering, singlelinecheck=false}
    \caption{Details of Training Dataset.}  
    \label{data_statistics_train_dataset} 
\end{figure*}
\section{Task Details}
\begin{figure*}[!htbp]  
    \centering
    \includegraphics[width=\textwidth]{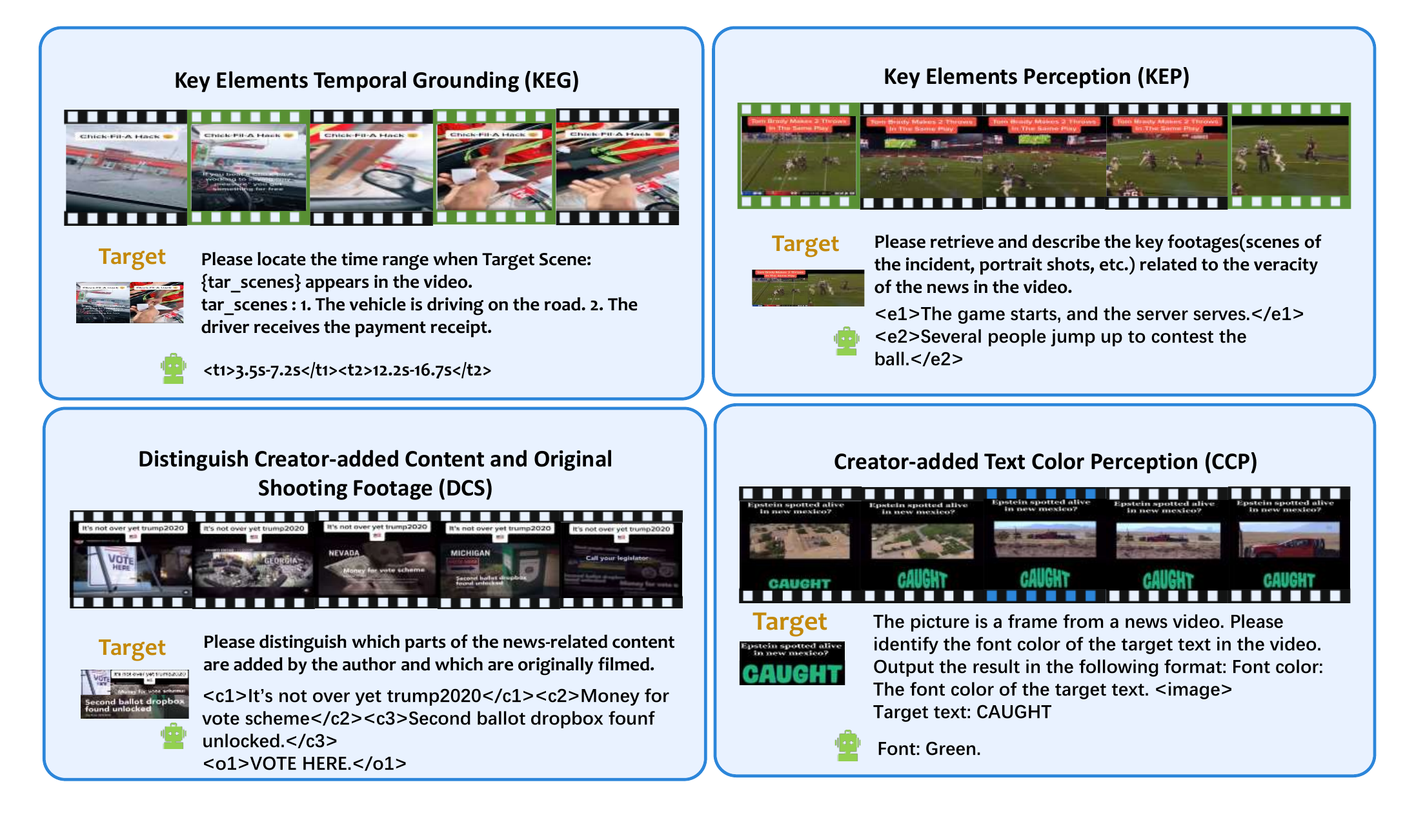}  
    \captionsetup{justification=centering, singlelinecheck=false}
    \caption{Definition of Evaluation Tasks}  
    \label{task_group1} 
\end{figure*}

\begin{figure*}[!htbp]  
    \centering
    \includegraphics[width=\textwidth]{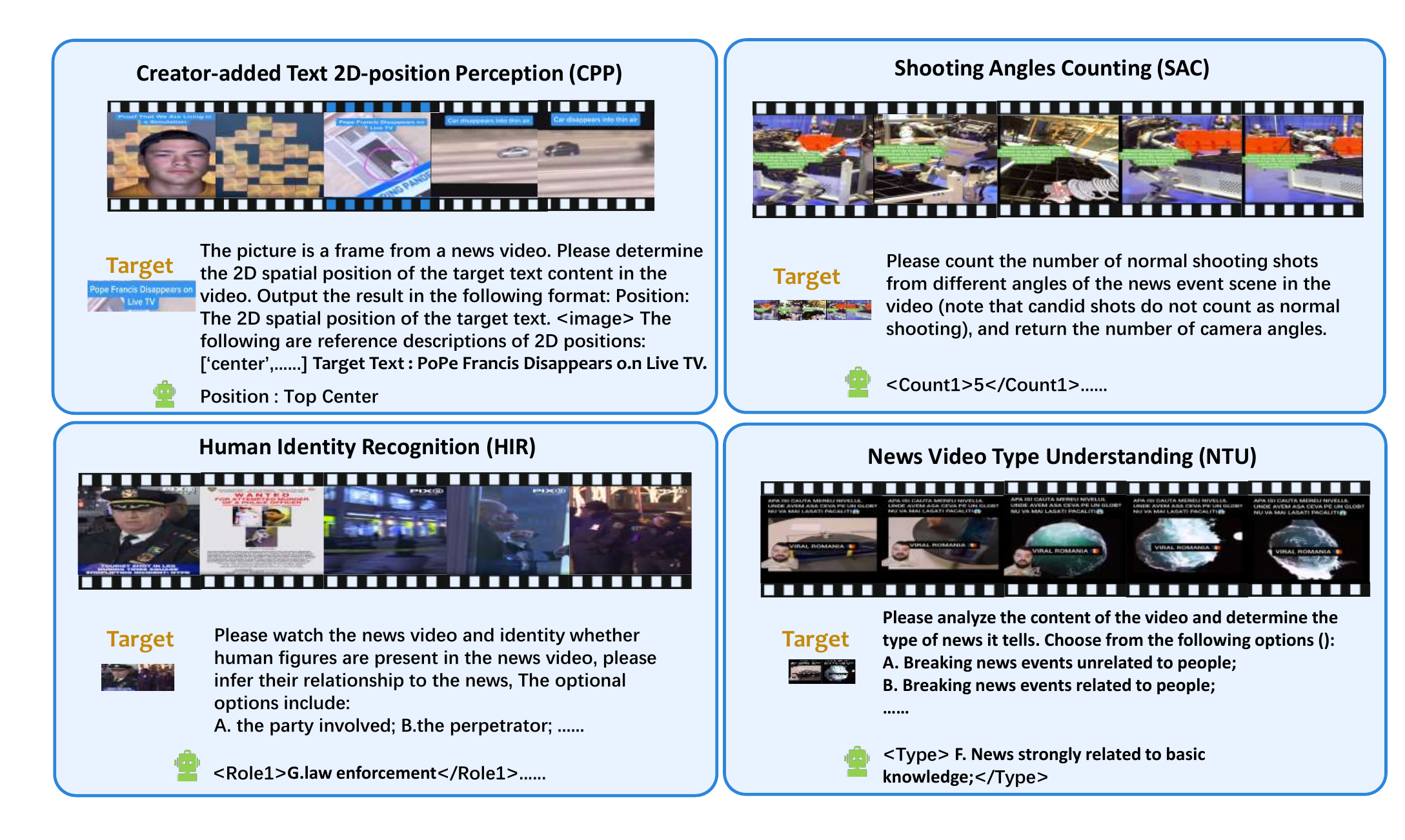}  
    \captionsetup{justification=centering, singlelinecheck=false}
    \caption{Definition of Evaluation Tasks}  
    \label{task_group2} 
\end{figure*}

\begin{figure*}[!htbp]  
    \centering
    \includegraphics[width=\textwidth]{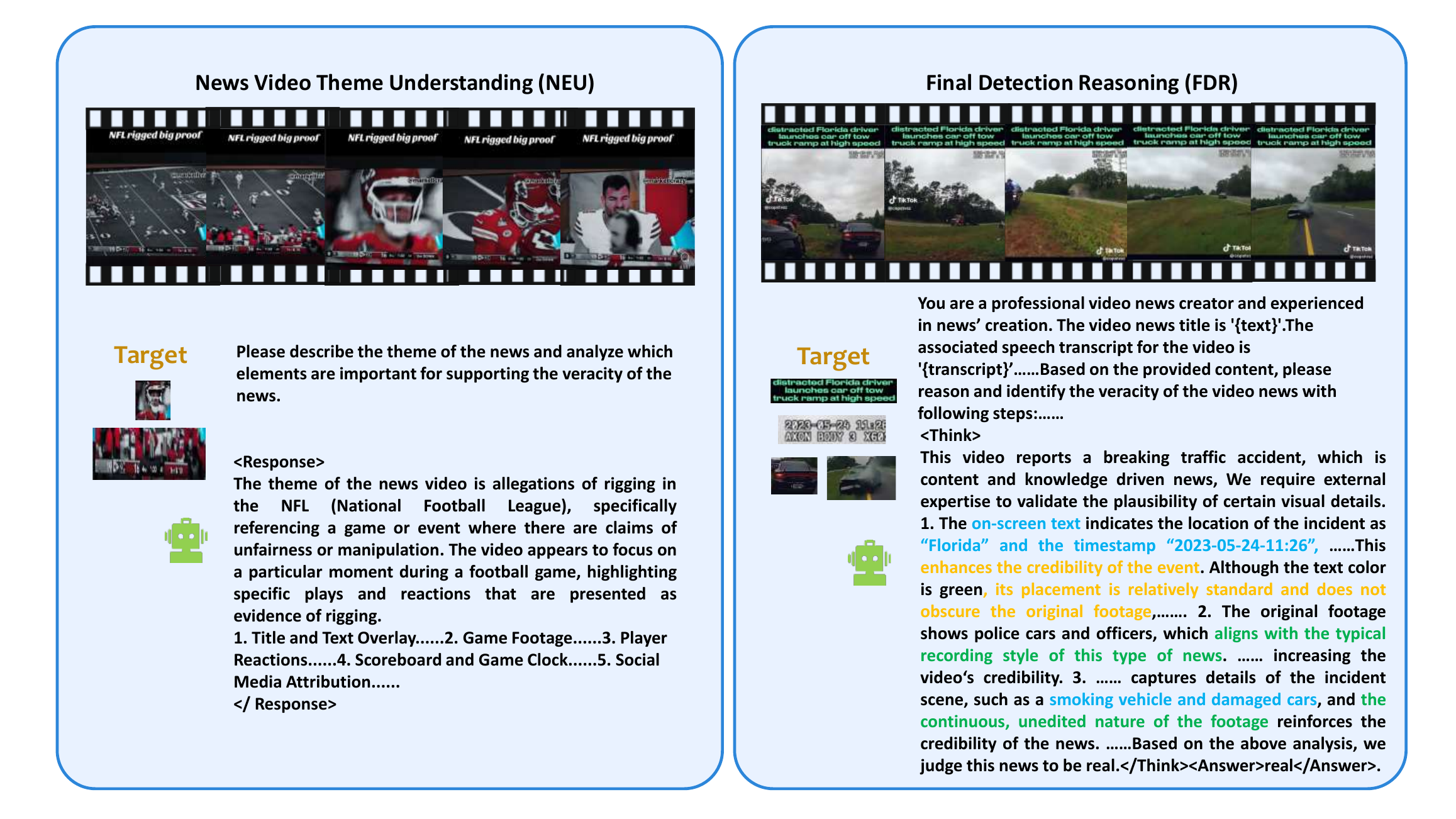}  
    \captionsetup{justification=raggedright, singlelinecheck=false}
    \caption{Definition of Evaluation Tasks. In the model output of the FDR process, the blue text represents entities identified in the video, including author-added text and original footage. The orange text denotes the rationales used during reasoning, comprising logical and empirical bases for determining authenticity. The green text indicates external knowledge related to the video employed during the reasoning process. These three components are used to evaluate the ENHR, KHR, and RHR of the reasoning process. The "Answer" is used to calculate the accuracy rate of the reasoning results.}  
    \label{task_group3} 
\end{figure*}

As demonstrated in Figure \ref{task_group1}, \ref{task_group2}, \ref{task_group3}. We define 10 tasks spanning perception, understanding, and reasoning, with diverse question formats including single selection, multiple selections, open-ended output, and structured output. In Key Elements Temporal Grounding (KEG), the model is instructed to identify the time range during which target elements appear, with each question containing at least one key element. Key Elements Perception (KEP) requires the model to retrieve key elements from video news and generate their semantic descriptions in a structured format. In Distinguish Creator-added Content and Original Shooting Footage (DCS), the model must recognize author-added content, such as on-screen text, and distinguish it from original shooting footage. Creator-added Text Color Perception (CCP) evaluates the model's ability to perceive the font color of target text. For Creator-added Text 2D-position Perception (CPP), the model identifies the screen position of target text, with semantic descriptions of all potential positions provided in the prompt to ensure accurate assessment. Shooting Angles Counting (SAC) requires the model to detect transitions between shooting angles and count their occurrences. In Human Identity Recognition (HIR), the model identifies all persons appearing in the video news through a multiple selection format. News Video Type Understanding (NTU) tasks the model with comprehending overall content to determine the news type from provided options. News Video Theme Understanding (NEU) requires the model to describe the global topic of the news video, including events, locations, and temporal information. Finally, Final Detection Reasoning (FDR) instructs the model to synthesize perceived elements, understood content, and reasoning rationales to assess the veracity of the news video.

\label{label_task_definitions}
\section{Data Statistics}

We annotate 2,871 training videos, each accompanied by visual information, semantic descriptions, and reasoning evidence. As shown in Figure \ref{data_statistics_train_dataset}, the training split contains an average of 4.58 key elements per video, totaling 8,233 elements. Key elements have an average duration of 13.59 seconds, and the number of key shots exhibits diverse distributions. The dataset maintains balanced representation across 6 news categories.

\section{Rationale Validation}
\label{label_rationale_validation}
During reasoning, MLLMs leverage their background knowledge and visual information from videos to generate interpretable, evidence-based outputs. These outputs consist of multiple rationales that collectively support the final detection judgment (fake or real). To validate the factual correctness of these rationales, we conduct a verification study on the original FakeSV dataset, examining whether the visual discrepancies described in the rationales actually exist in the source videos. Beyond validation, we also provide practical explanations for identified discrepancies. We categorize rationales into two types: those relevant to creator-added content (CAC) and those relevant to original shooting footage (OSF). For CAC rationales validation, we employ the automated pipeline illustrated in Figure \ref{creator-added_text_sta_process} to extract visual features. For OSF rationales validation, we leverage the expert annotations in POVFNDB for validation. Our empirical analysis reveals that the distinguishing characteristics between fake and real news described in MLLM-generated rationales are corroborated by statistical patterns observed in the original FakeSV dataset. This empirical validation enhances the reliability of the MLLM reasoning process.

\subsection{CAC Features Extraction}
\begin{figure*}[!htbp]  
    \centering
    \includegraphics[width=\textwidth]{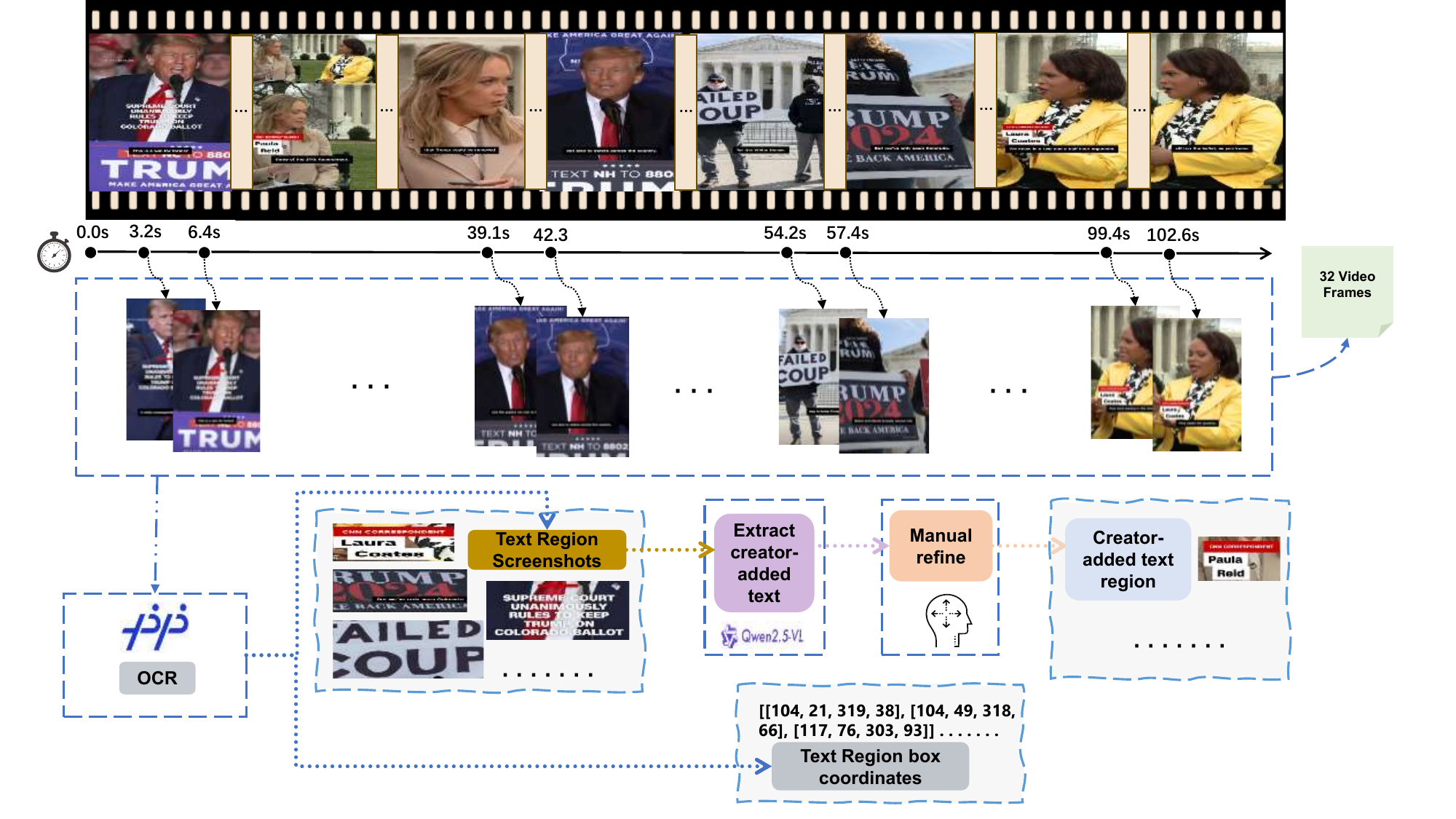}  
    \captionsetup{justification=centering, singlelinecheck=false}
    \caption{Creator-added Content Extraction Process}  
    \label{creator-added_text_sta_process} 
\end{figure*}

\begin{figure}[t]
    \centering
    \begin{minipage}{0.22\textwidth}
        \centering
        \includegraphics[height=5cm, keepaspectratio]{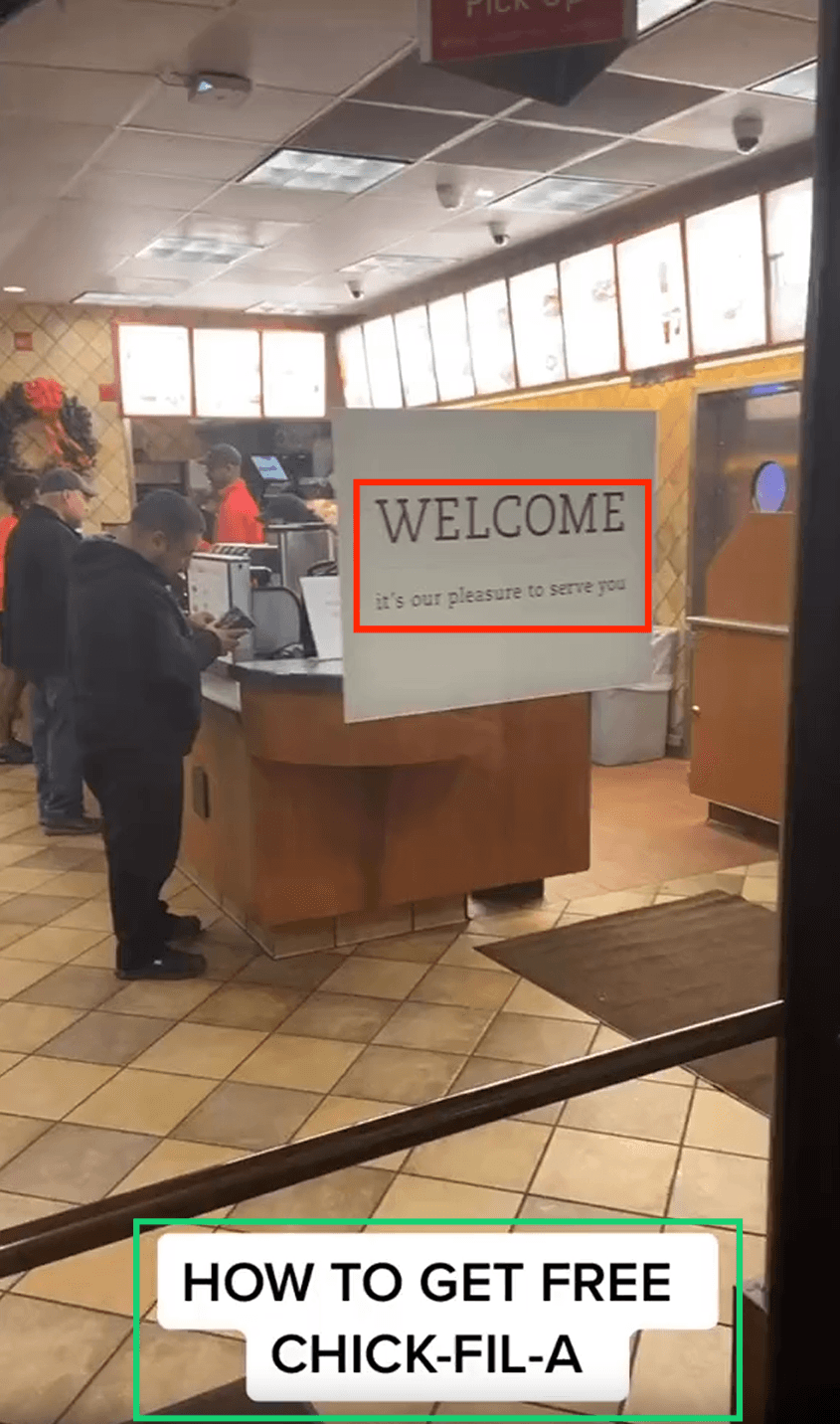}
        \captionof{figure}{CAC in green box,OSF in red box.}
        \label{text-region}
    \end{minipage}
    \hfill
    \begin{minipage}{0.22\textwidth}
        \centering
        \includegraphics[height=5cm, keepaspectratio]{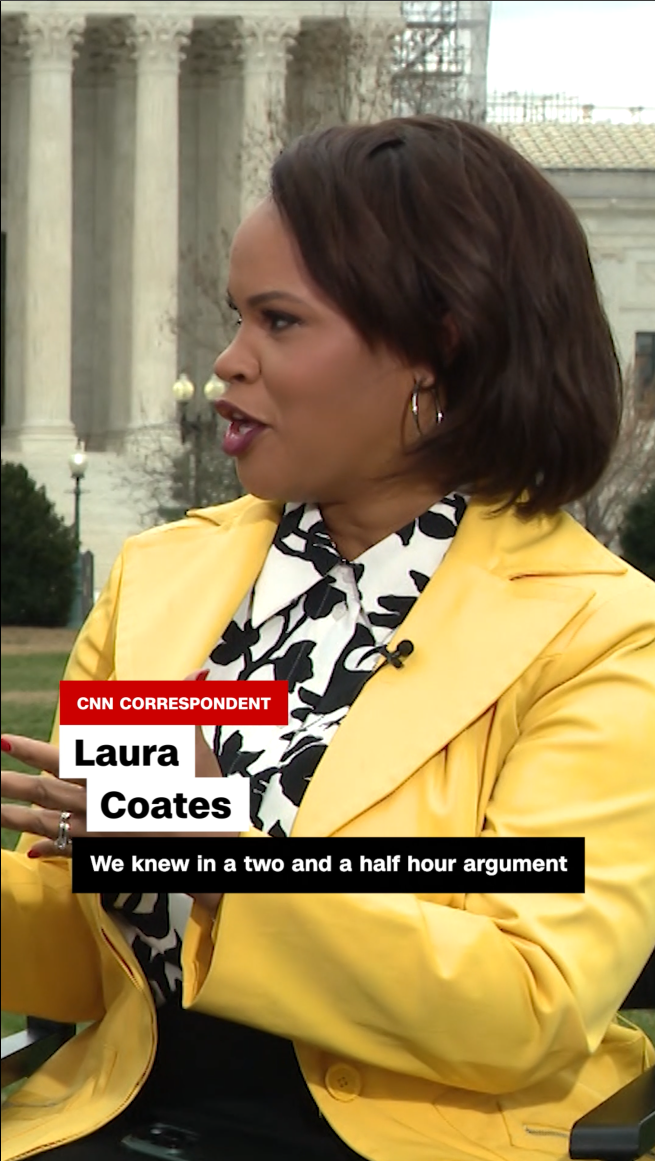}
        \captionof{figure}{Individual with identity}
        \label{individual-with-identity}
    \end{minipage}
\end{figure}

Creator-added content (CAC) in news videos predominantly consists of textual overlays inserted using specialized editing software. These overlays serve various functions, including providing event context through captions, introducing key figures, and displaying distinctive graphical identifiers from authoritative media outlets.
To enable quantitative analysis of CAC rationales, we develop an automated extraction pipeline as shown in Figure \ref{creator-added_text_sta_process}. First, we uniformly sample 16 frames per video using FFmpeg with temporally equidistant intervals. Second, we apply PaddleOCR to detect text regions in each frame, obtaining bounding boxes and region screenshots. Since PaddleOCR occasionally detects text from the original shooting footage (e.g., signs shown in Figure \ref{text-region}), we manually filter such instances to retain only creator-added text regions. Third, we employ Qwen2.5-VL-72B with designed prompt templates to extract semantic attributes of each text region, including content, color, 2D position, and aspect ratio. Finally, domain experts refine the MLLM-generated descriptions to ensure accuracy.

\subsection{CAC Rationales Validation}

Creator-added text is typically used in video news to highlight key information, such as sites, date. We select two rationales about text font color, screen position, emotion level are frequently cited by MLLMs. We utilize visual information generated by the pipeline in the Figure \ref{creator-added_text_sta_process}. We employee the PaddleOCR tools to extract text regions from video and process them with the pipeline in Figure \ref{creator-added_text_sta_process}.

\begin{figure}[t]  
    \hspace{-15pt}
    \includegraphics[scale=0.49]{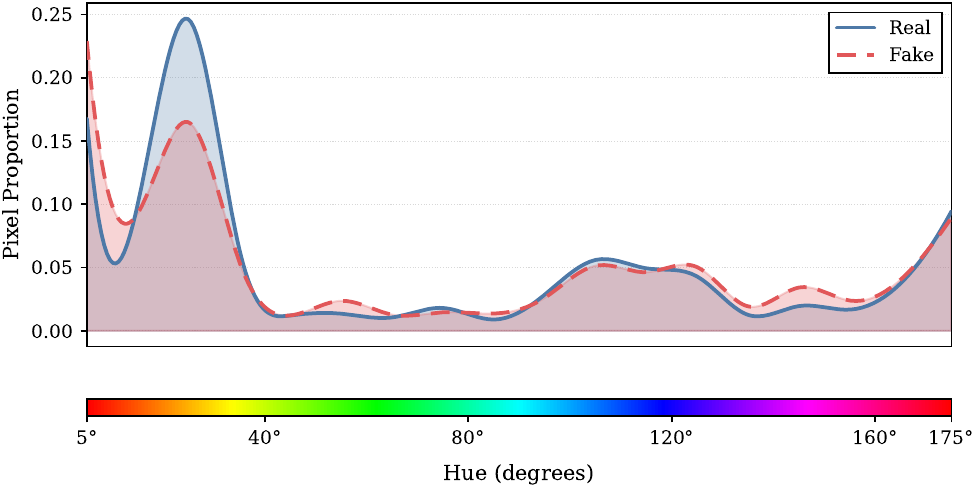}  
    \captionsetup{justification=centering, singlelinecheck=false}
    \caption{Hue distribution in creator-added content \\ of fake/real news video }  
    \label{hue_distribution}
\end{figure}
\subsubsection{Color Distribution}
\textbf{Rationale-1.} Yellow is a professional typographic color choice for video news content, offering high visibility and prominence while maintaining a neutral-to-bright aesthetic. This color selection minimizes emotional bias in viewers' information processing and enhances the perceived authenticity of news content.\\
\textbf{Rationale-2.} Although red typography achieves high visual salience, it induces emotional arousal in audiences, which contradicts the factual nature of news dissemination and consequently diminishes perceived authenticity.\\
\textbf{Analysis.} Prior work has shown that font color affects news comprehension, with hue serving as the primary color identifier \cite{zhou2022effects}. To investigate whether color usage differs between real and fake news, we conduct a statistical analysis comparing pixel-level hue distributions in creator-added text across both categories. As shown in Figure \ref{hue_distribution}, we observe significant divergence in hue distribution between real and fake news videos, particularly in red and yellow ranges. This finding aligns with MLLM-generated rationales that frequently cite font color as a veracity indicator.\\
We attribute this phenomenon to differences in production practices and intent. Fake news producers tend to employ emotionally salient colors (particularly red) to manipulate audience perception and enhance credibility of false claims. In contrast, creators of authentic news use formal, neutral color schemes to guide viewer attention to substantive information. Furthermore, the prevalence of high-saturation red in fake news may reflect rushed production workflows, where creators prioritize immediate visual impact over context-appropriate color selection. This production haste results in overreliance on attention-grabbing colors without consideration for content alignment.
\begin{figure}[htbp]
    \centering  
    
    \begin{subfigure}[b]{0.23\textwidth}
        \centering
        \includegraphics[width=\linewidth]{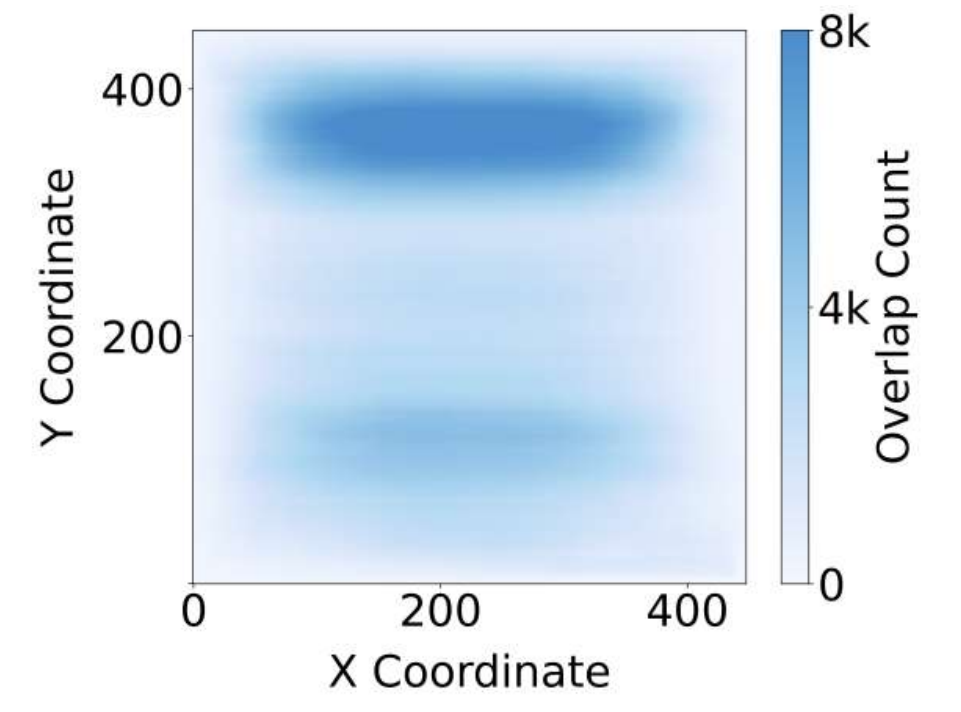}  
        \caption{Distribution of text region in fake news video} 
        \label{subfig:fake}  
    \end{subfigure}
    \hfill  
    \begin{subfigure}[b]{0.23\textwidth}
        \centering
        \includegraphics[width=\linewidth]{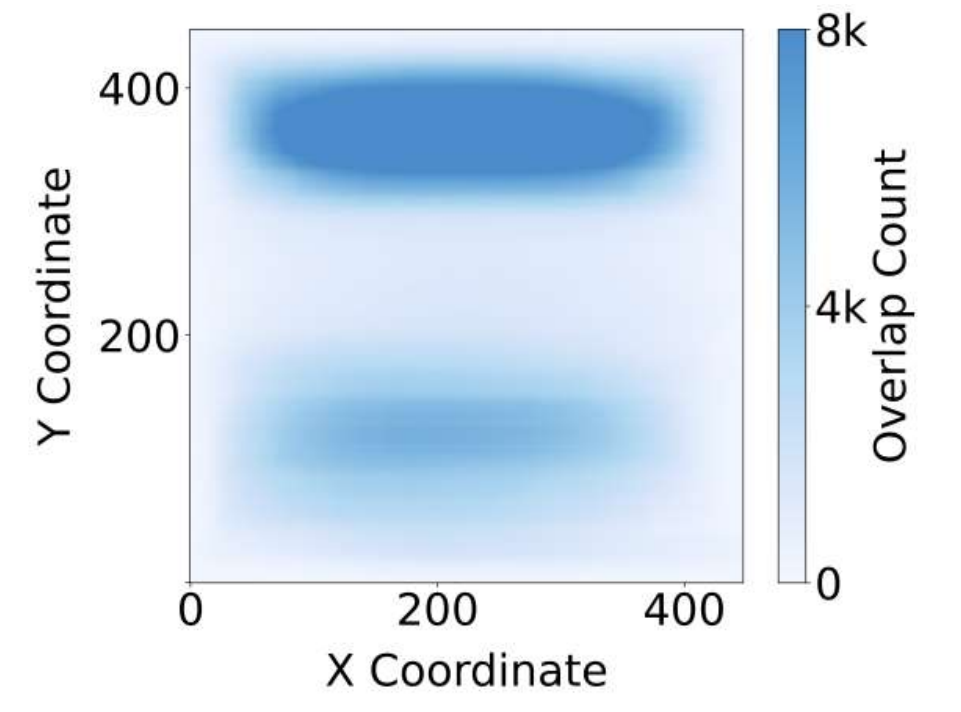}  
        \caption{Distribution of text region in real news video}  
        \label{subfig:real}  
    \end{subfigure}
    
    \caption{Comparison of heatmaps}
    \label{heatmap_compare}  
    \vspace{-12pt} 
\end{figure}

\subsubsection{Spatial Distribution}
\textbf{Rationale-1.} The text describing the event location and time is positioned at the top-center of the screen, consistent with conventional placement patterns in real news videos.\\
\textbf{Rationale-2.} Text positioned at the screen center occludes the original footage, severely interfering with viewers' access to authentic visual content. This layout pattern is commonly employed in fake news to obscure factual evidence, thereby reducing perceived authenticity.\\
\textbf{Analysis.} As illustrated in Figure \ref{heatmap_compare}, the spatial distribution of text regions exhibits notable differences between real and fake news videos, consistent with patterns described in MLLM-generated rationales.
We attribute the observed distributional discrepancy to two factors.1) Fake news producers strategically overlay text on video footage to obscure factual details in the original visuals and redirect viewer attention toward fabricated textual narratives. 2) Fake news creators lack professional training in screen-text composition. Unlike authentic news production, where text placement follows established design principles that balance information salience with minimal occlusion of original content, fake news exhibits arbitrary positioning that disregards these compositional best practices. 

\subsection{Original Shooting Footage}
Original shooting footage (OSF) typically provides first-hand documentation of events, particularly for breaking incidents. Unlike CAC, OSF is less susceptible to manipulation, thereby offering higher credibility. To investigate whether OSF-related features correlate with news veracity, we select rationales from MLLM reasoning that reference OSF characteristics and analyze their distribution across real and fake news videos using annotations from POVFNDB.
\subsubsection{Key footage Distribution}
\begin{figure}[t]
  \includegraphics[width=\columnwidth]{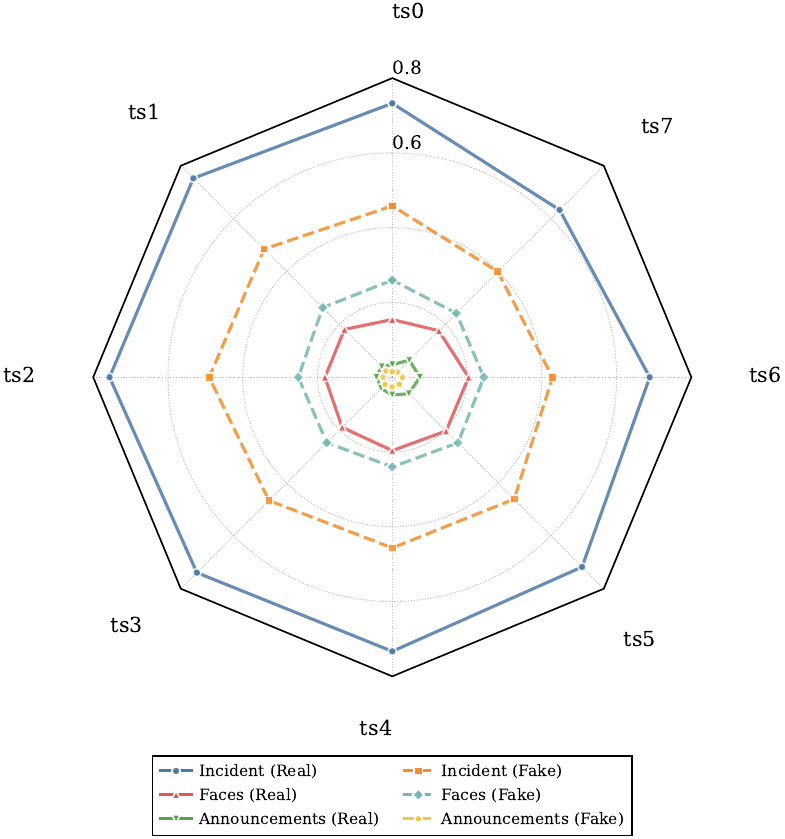}
  \captionsetup{justification=raggedright, singlelinecheck=false}
  \caption{The proportion of key footage appearing in different time segments of fake and real news videos. Here, "on\_site" refers to footage from the event scene, "close\_up" indicates close-up shots of individuals related to the event, and "declaration" represents statements from authoritative institutions regarding the event. "ts0-ts7" denotes segments of the video divided uniformly along the timeline, where "ts0" refers to the first segment of the video, "ts7" refers to the final segment, and so forth.}
  \label{time_range_distribution}
\end{figure}
\textbf{Rationale-1} The video concludes with screenshots of official police documents detailing the incident and involved parties, presenting authoritative information to viewers. This placement pattern—authoritative evidence at video end—is characteristic of authentic news reports and enhances perceived credibility.\\
\textbf{Rationale-2} The video reports a breaking news event but lacks on-site footage, instead presenting multiple unrelated scenes. This absence of relevant visual evidence reduces perceived authenticity.\\
\textbf{Analysis} Key frames are closely tied to news themes and hold significant evidentiary value for veracity detection. In professional news production, critical information is typically positioned in opening segments to engage viewers, while supplementary content such as disclaimers and announcements appears in closing segments.
To examine temporal distribution patterns, we analyze three types of salient visual content: event scenes, human faces, and textual statements. We uniformly partition each video into eight temporal segments and employ Qwen2.5-VL-72B to detect the presence of these three frame types in each segment. Figure \ref{time_range_distribution} illustrates the resulting distributions, which align with patterns described in MLLM-generated rationales. We attribute the observed differences between real and fake news to two factors: fake news producers' deliberate attempts to obscure factual evidence and their lack of professional training in news composition conventions.
\subsubsection{Subject Identity Distribution}
\begin{figure}[t]
  \includegraphics[width=\columnwidth]{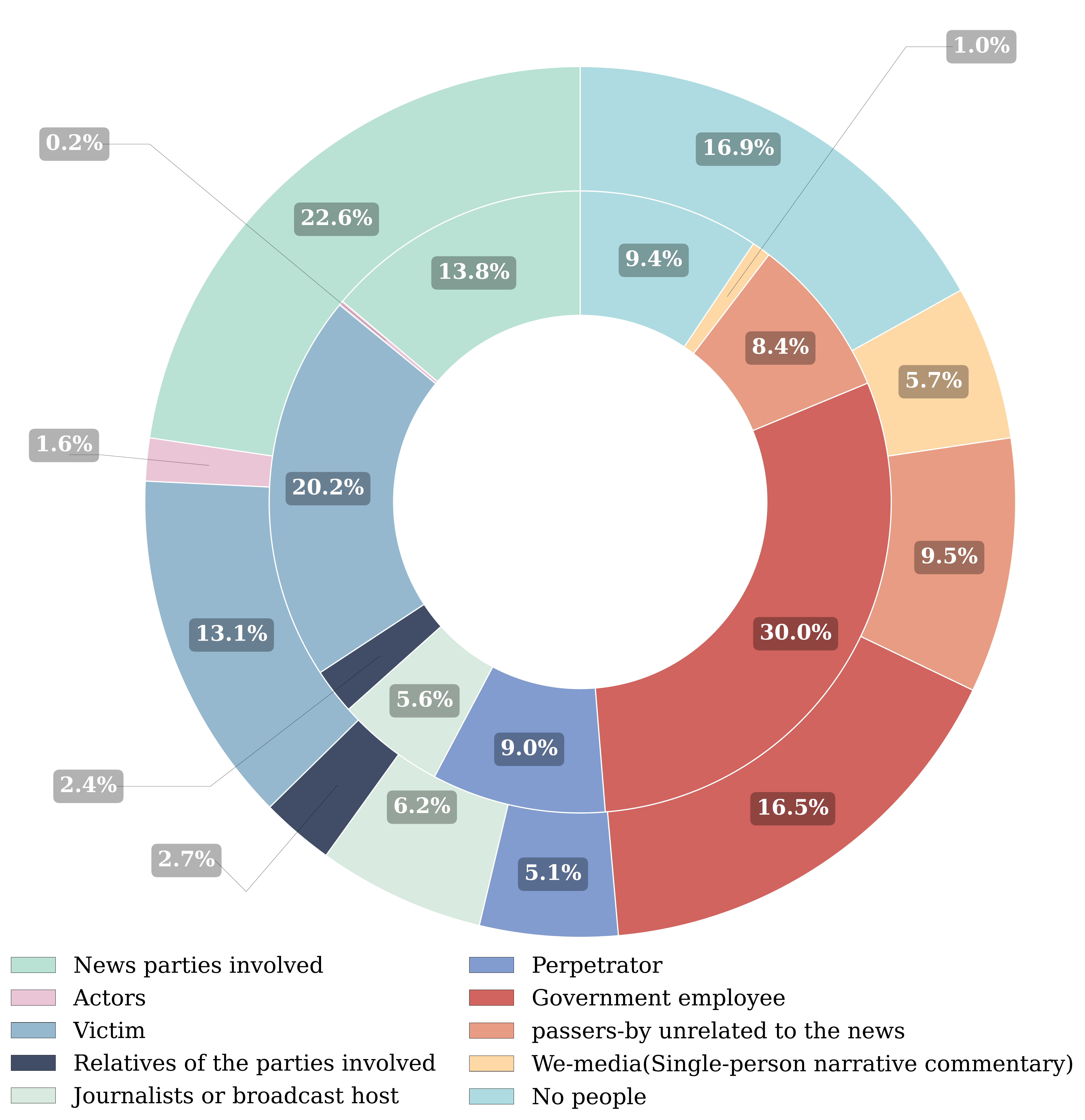}
  \caption{Distribution of subject identity in\\ fake(external)/real(internal) news video.}
  \label{selections_distribution_detain_label}
\end{figure}
\textbf{Rationale-1} The video features interviews with law enforcement personnel, with names displayed via on-screen captions. This practice is commonly employed in incident reporting to enhance news credibility.\\
\textbf{Rationale-2} Despite being framed as breaking incident news, the video contains no appearances by event-related individuals, substantially reducing credibility.\\
\textbf{Rationale-3} The video documents an incident and includes footage of the victim receiving medical treatment, with on-screen name identification. This verifiable content substantially enhances perceived news authenticity.\\
\textbf{Analysis} News videos frequently feature human subjects to convey key event information through specific visual formats, such as face-to-face interviews and press conference footage. These individuals include involved parties, victims, law enforcement personnel, and self-media creators, among others. However, the reliability of information provided by different identity types varies significantly \cite{doi:10.1177/13684302211030004}. To investigate this phenomenon, we analyze the proportional distribution of identity types across entire video durations.
Figure \ref{selections_distribution_detain_label} reveals distinct distributional patterns between real and fake news. Perpetrators, victims, and official personnel appear more frequently in real news videos, with law enforcement officers appearing more than twice as often in real news compared to fake news. Notably, 22.6
We attribute these disparities to two primary factors. First, real news producers possess institutional access to law enforcement personnel involved in reported events, whereas fake news creators lack such credentials. Second, the inclusion of authoritative figures serves as an essential verification mechanism to mitigate audience skepticism in authentic news production—an element that fake news producers deliberately avoid. Conversely, fake news videos exhibit higher proportions of both absent human subjects and self-media creators, reflecting their reliance on unofficial sources and reduced evidentiary standards.

\begin{figure}[t]
  \includegraphics[width=\columnwidth]{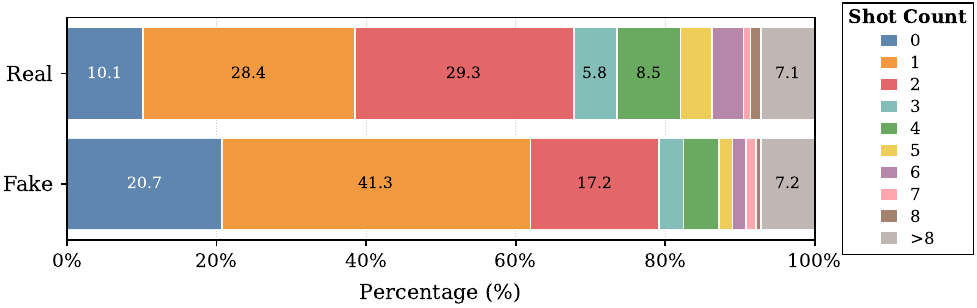}
  \caption{Number of shooting angles in the fake and real video news}
  \label{final_key_shots_distribution}
\end{figure}

\subsubsection{Relevant shooting angles}
\textbf{Rationale-1} The video captures the event from multiple perspectives, enabling viewers to comprehend the complete context and thereby enhancing perceived veracity.\\
\textbf{Rationale-2} The video contains only a single static image of the event scene, preventing cross-verification of news content and reducing perceived authenticity.\\
\textbf{Analysis} We leverage annotated data from the benchmark to analyze the distribution of camera shots between fake and real news videos. As shown in Figure \ref{final_key_shots_distribution}, fake news exhibits a higher proportion of videos with 0-1 shots, whereas real news dominates in videos with more than one shot. These patterns align with MLLM-generated rationales.
To provide comprehensive and objective coverage of news events, professional creators employ multiple shooting angles within their videos. This approach enables audiences to gain a thorough understanding of event progression while enhancing the credibility and persuasiveness of news reports \cite{doi:10.1177/19312431231157104}. We attribute the observed disparities to three primary factors. First, fake news producers typically lack sufficient shooting materials about the reported events, limiting their ability to present multiple perspectives. Second, reduced shot diversity prevents audiences from fully comprehending the facts, which paradoxically serves the obfuscation goals of disinformation creators. Third, fake news producers often lack the professional videography skills required for seamless camera shot transitions in news production.


\section{Additional Experiments}
We evaluated the performance of MLLMs across various capabilities on the POVFND task, utilizing both process-based and result-based tasks within the benchmark.We further verify the effectiveness of different visual prompts by comparing three variants: CAC-POVFND-CoT, OSF-POVFND-CoT, and ALL-POVFND-CoT, finding that OSF yields the most substantial performance gains. Through further analysis of the evaluation process, we identified several insights and discussed their underlying intrinsic causes.
\begin{table*}[t] 
    \centering 
    \resizebox{\linewidth}{!}{  
        \begin{tabular}{lcccccccccc}
            \toprule
            \textbf{Model} & \multicolumn{2}{c}{\textbf{CAC-POVFND-CoT}} & \multicolumn{2}{c}{\textbf{OSF-POVFND-CoT}} & \multicolumn{2}{c}{\textbf{ALL-POVFND-CoT}} & \multicolumn{2}{c}{\textbf{POVFND-CoT}} & \multicolumn{2}{c}{\textbf{Instruct-Tuning}} \\
            \cmidrule(lr){2-3} \cmidrule(lr){4-5} \cmidrule(lr){6-7} \cmidrule(lr){8-9} \cmidrule(lr){10-11}  
            & ACC & M-F1 & ACC & M-F1 & ACC & M-F1 & ACC & M-F1 & ACC & M-F1 \\ 
            \midrule
            Gemini-2.5-flash    & \best{78.88} & \best{78.81} & \best{79.02} & \best{78.87} & \best{78.75} & \best{79.66} & \best{78.61} & \best{77.87} & - & - \\
            GPT-4o-mini         & 72.24 & 73.56 & 72.00 & 73.45 & 71.31 & 74.66 & 71.05 & 72.35 & - & - \\
            Qwen2.5VL-72b       & \secondbest{77.95} & 74.08 & \secondbest{78.35} & 74.19 & \secondbest{77.56} & \secondbest{78.12} & \secondbest{77.00} & \secondbest{76.05} & - & - \\
            Qwen2.5VL-32b       & 74.90 & 69.16 & 74.76 & 74.68 & 75.43 & 70.69 & 74.90 & 74.23 & - & - \\
            Qwen2.5VL-7b        & 69.85 & 65.66 & 70.38 & 71.77 & 70.25 & 69.75 & 70.12 & 71.51 & - & - \\
            InternVL3-78b       & 76.49 & \secondbest{77.04} & 76.89 & \secondbest{75.58} & 76.62 & 74.15 & 75.96 & 72.45 & - & - \\
            InternVL3-38b       & 73.43 & 70.64 & 73.97 & 70.66 & 73.71 & 71.56 & 73.43 & 71.65 & - & - \\
            InternVL3-8b        & 70.51 & 69.14 & 70.39 & 68.51 & 70.65 & 69.16 & 70.00 & 71.55 & - & - \\
            Qwen2.5VL-7b-Instruct & 57.23 & 53.71 & 57.50 & 55.39 & 57.76 & 55.36 & 58.43 & 58.23 & \best{81.14} & \best{80.26} \\
            \bottomrule
        \end{tabular}
    }
    \caption{Evaluation results of different CoT frameworks on fake news detection. CAC-POVFND-CoT integrates creator-added content analysis (color, position, text), OSF-POVFND-CoT incorporates original shooting footage analysis (key elements, shooting angles, human identities), and ALL-POVFND-CoT combines both CAC and OSF components with CoT reasoning.}
    \label{cot_reason_result_detail}
    \vspace{-12pt}
\end{table*}

\begin{figure*}[t]
  \centering
  \includegraphics[width=\textwidth]{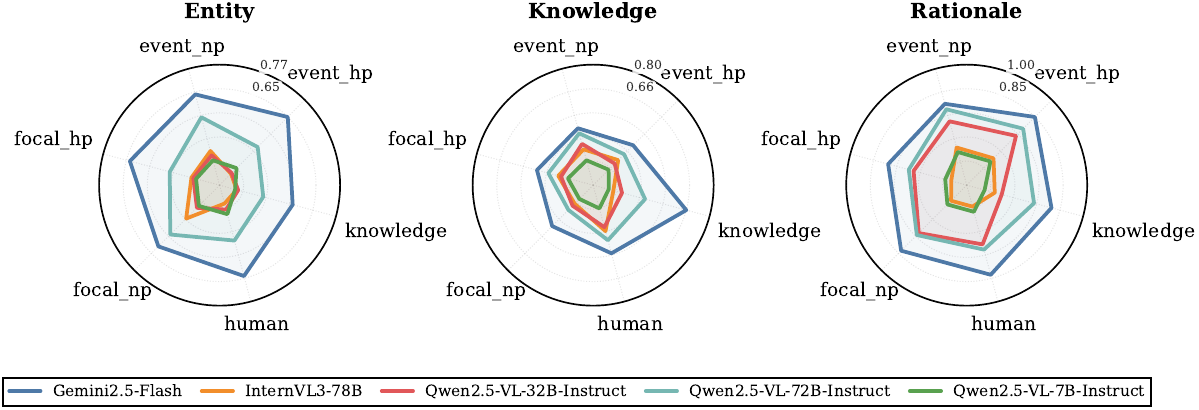}  
  \captionsetup{justification=centering, singlelinecheck=false}
  \caption{Comparison of entity hit rate, knowledge hit rate and rationale hit rate in reasoning output.}
  \label{combined_radar_reason_entity_knowledge_rationale}
\end{figure*}
\begin{figure}[t]
  \includegraphics[width=\columnwidth]{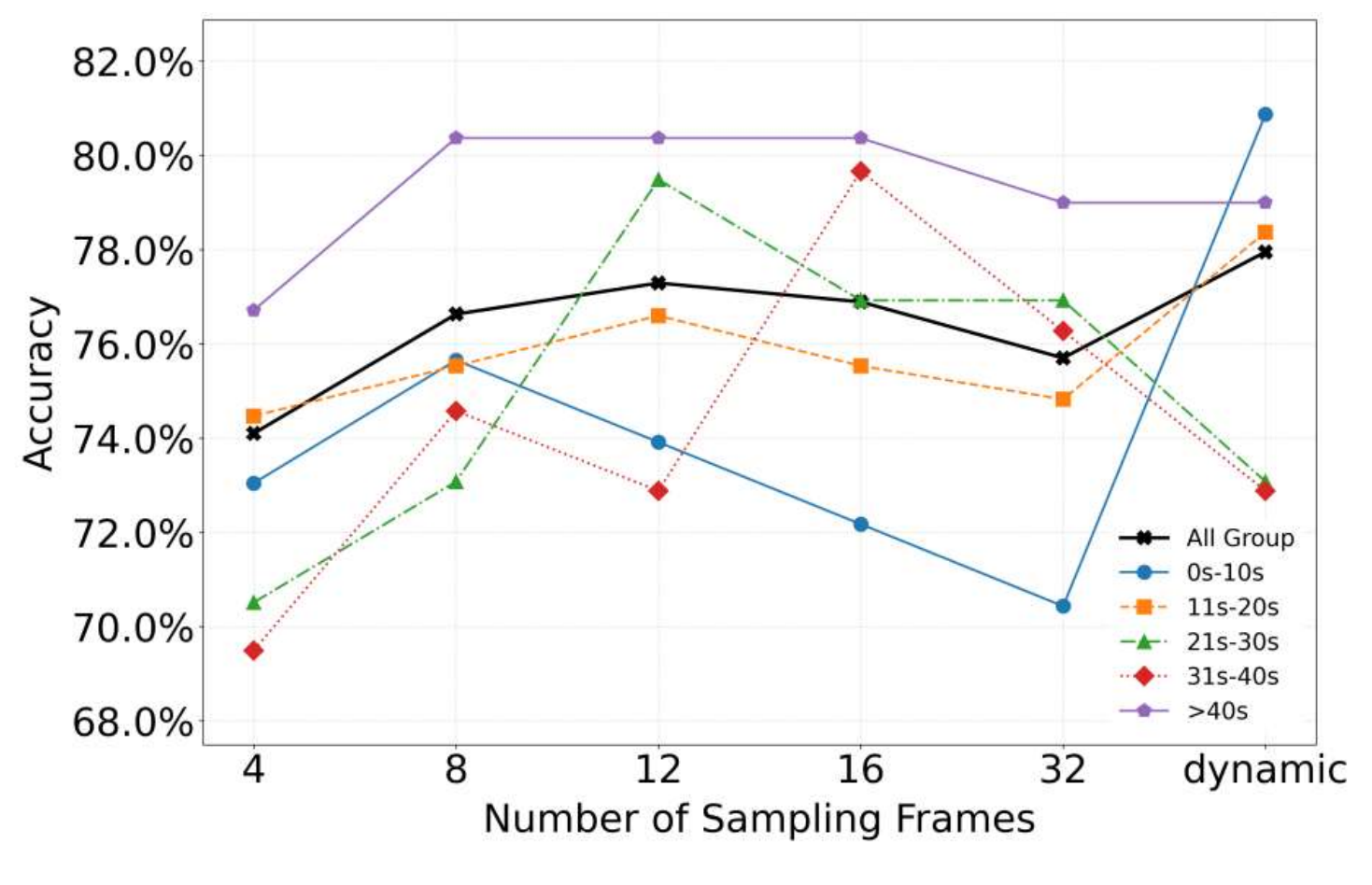}
  \caption{Comparison of Accuracy with Different Number of Sampling Frames and Duration of News Video. Dynamic Sampling FPS=1 and the Max Number of Frames is 16.}
  \label{duration_group_acc_trend_analysis_final}
\end{figure}
\begin{figure}[t]
  \includegraphics[width=\columnwidth]{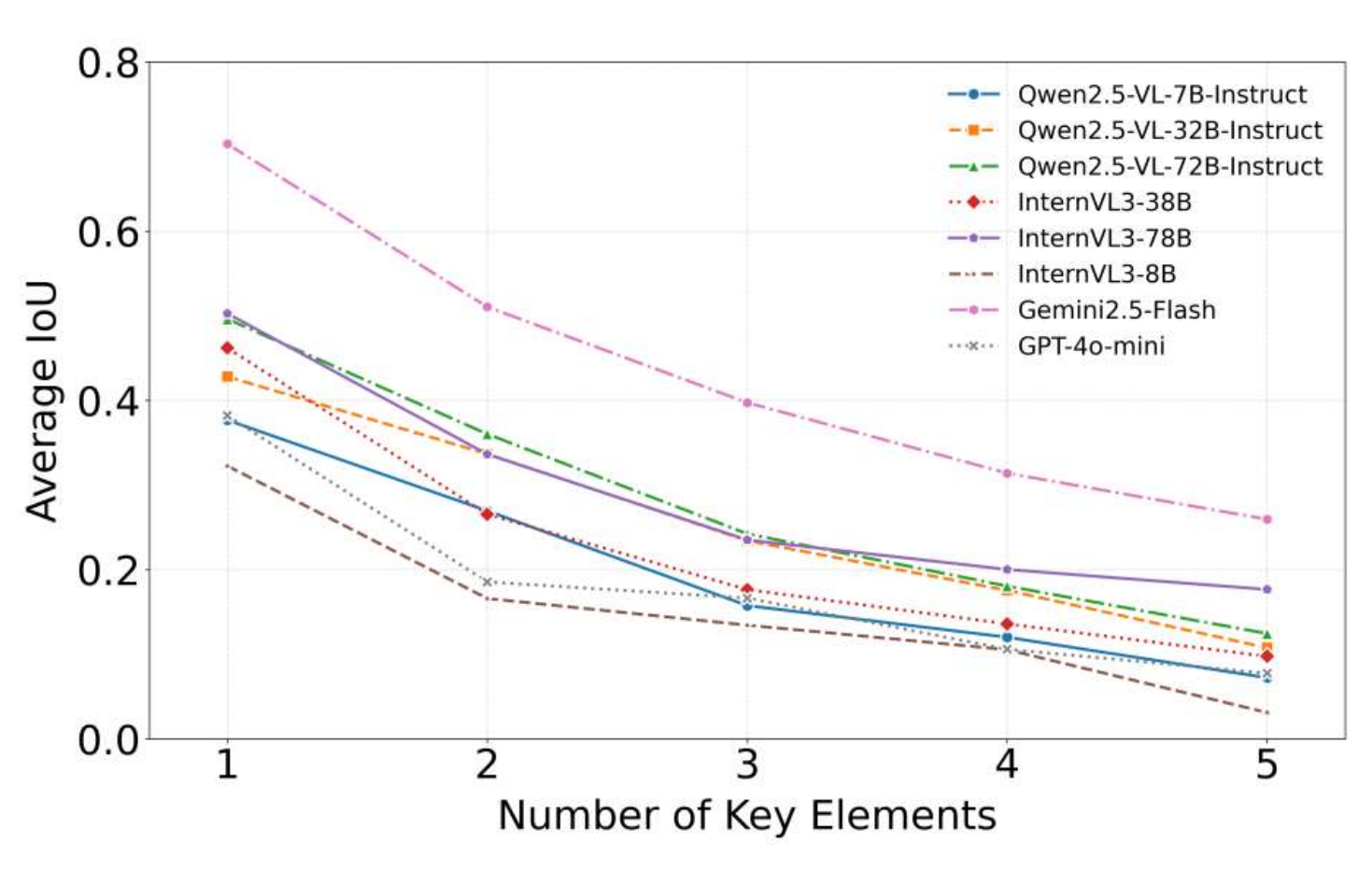}
  \caption{Comparison of IoU with Different Number of Key elements in the Video}
  \label{key_elements_IoU_plot}
\end{figure}
\begin{figure}[t]
  \includegraphics[width=\columnwidth]{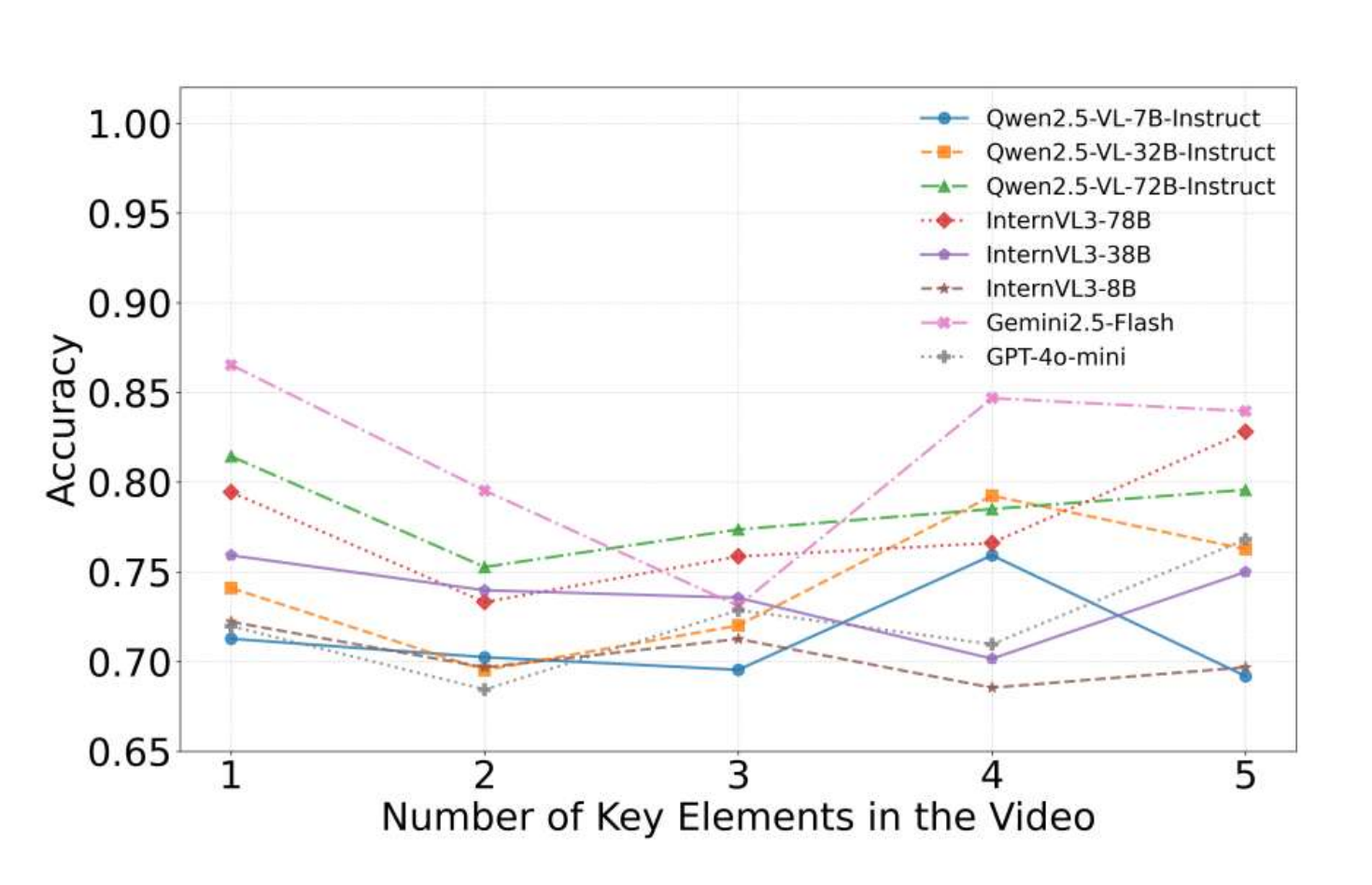}
  \caption{Comparison of Accuracy with Different Number of Key elements in the Video}
  \label{model_key_elements_num_accuracy_chart}
\end{figure}

\textbf{Insight 1. For POVFND task, the performance bottlenecks of the model vary across different types of news videos.} As presented in Figure \ref{combined_radar_reason_entity_knowledge_rationale}, reasoning relies on different information types across distinct news video categories. It can be observed that knowledge-based news videos rely more heavily on knowledge during the reasoning process, while exhibiting low utilization of entities in the video (e.g., screen text and footage). In contrast, the event-based news videos are less sensitive to external knowledge, and they utilize far more entities derived from the video contents. \\
\textbf{Discussion 1.} We attribute this bias to the fact that knowledge-based news, false information is typically embedded in the video scripts, whereas the video contents mostly are irrelevant to the authenticity of the news. Instead, the factual claims and logical relationships within the scripts constitute the primary targets for verification; the verification for veracity of event-based news typically focuses on determining whether the target event occurred. This process relies on the complete recording of the target event in the video; otherwise, the subsequent reasoning process lacks a valid evidence.\\
\textbf{Insight 2. For POVFND task, dense frame sampling is not requisite for video content perception. Video frame sampling strategies should take into account both video duration and model input load.} As shown in Figure \ref{duration_group_acc_trend_analysis_final}, we take the performance of InternVL3-78B as an example. For each video duration group, there exists an optimal number of sampling frames. And this optimal number tends to increase as the video duration extends. But dense frame sampling does not guarantee a continuous improvement in accuracy; instead, it may even compromise the model performance. Dynamic frame sampling strategy yields higher accuracy, especially for shorter news videos.\\
\textbf{Discussion 2.} We ascribe the result to the news videos are typically short in duration, with key information scattered and numerous redundant frames. For authenticity detection, global perception is more important but fine-grained. Once the number of sampled frames is sufficient for the model to capture the video's key information, the bottleneck of accuracy will shift from the model's multi-modal perception capability to its FND reasoning capability. Continuing to increase the number of frames at this point will only impose greater input pressure on the model, resulting in performance degradation.\\
\textbf{Insight 3. For POVFND task, better performance can only be achieved with sufficient key elements in the video and the model's temporal localization capability.} As demonstrated in Figure \ref{model_key_elements_num_accuracy_chart}, when the number of key elements is one, all models achieve relatively favorable performance. But as the number of key elements in the video increases, only some more advanced models exhibit improved performance.\\
\textbf{Discussion 3.} We attribute this phenomenon to the fact that, the number of key elements generally increases as the news video duration grows. Thus, MLLMs can handle most samples containing only one element. However, as the number of elements increases, less capable models fail to capture the key information within these longer videos. As presented in Figure \ref{key_elements_IoU_plot}, model's temporal localization capability decrease with the number of key elements needed to be localized. However, MLLMs that capture key information can obtain more descriptions of news content, thereby improving their accuracy.

\textbf{Insight 4. Visual prompts require adaptive MLLM capacities.} As shown in Table \ref{cot_reason_result_detail}, combining CoT with CAC or OSF features improves performance in more advanced models (Gemini-2.5-Flash and Qwen2.5-VL-72B), but these gains diminish in smaller-scale MLLMs.
\textbf{Discussion 4.} We attribute this to smaller models' limited ability to utilize critical information in prompts, causing features to become noise that degrades performance. This indicates that effective visual information integration for VFND depends on MLLMs' adaptive visual processing capabilities.\\

\section{Addition Training Details}
\begin{table}[t]
\centering
\small
\begin{tabular}{ll}
\toprule
\textbf{Parameter} & \textbf{Value} \\
\midrule
\multicolumn{2}{l}{\textit{Model Configuration}} \\
Base Model & Qwen2.5-VL-7B-Instruct \\
Fine-tuning Method & LoRA \\
Template & qwen2\_vl \\
Flash Attention & Auto \\
\midrule
\multicolumn{2}{l}{\textit{Training Setup}} \\
Dataset & all\_povfndb\_reasoning\_train \\
Max Samples & 100,000 \\
Sequence Length & 32,768 \\
Number of Epochs & 12 \\
Batch Size per Device & 2 \\
Gradient Accumulation Steps & 2 \\
Effective Batch Size & 32 \\
\midrule
\multicolumn{2}{l}{\textit{Optimization}} \\
Optimizer & AdamW \\
Learning Rate & 5e-5 \\
LR Scheduler & Cosine \\
Warmup Steps & 100 \\
Max Gradient Norm & 1.0 \\
Precision & BF16 \\
\midrule
\multicolumn{2}{l}{\textit{Vision-Language Settings}} \\
Freeze Vision Tower & True \\
Freeze MM Projector & False \\
Freeze Language Model & False \\
Image Max Pixels & 50,176 \\
Image Min Pixels & 10,000 \\
Video Max Pixels & 65,536 \\
Video Min Pixels & 256 \\
\bottomrule
\end{tabular}
\caption{Training hyperparameters and configurations for the Qwen2.5-VL-7B-Instruct model fine-tuning. all\_povfndb\_reasoning\_train is generated by ALL-POVFNDB-CoT reasoning process.}
\label{training_params}
\end{table}
\section{Metrics Detail}
\label{appendix_lab_metrics}
To evaluate MLLM performance across the 10 tasks, we define task-specific metrics tailored to output formats and evaluation objectives. This section details the metric definitions, calculation procedures, and relevant datasets. \\
\textbf{Avg.EHR (Element Hit Rate):} 
EHR measures the proportion of ground-truth elements correctly identified in model predictions. 
For a given video $j$, let the prediction be $\hat{y}_j = \{\hat{e}_1, \hat{e}_2, \ldots, \hat{e}_n\}$ and the ground-truth label be $y_j = \{e_1, e_2, \ldots, e_m\}$, where $n$ and $m$ denote the number of predicted and ground-truth elements, respectively.

To evaluate each ground-truth element $e_i$, we employ GPT-4 as a semantic matching function $M(\cdot, \cdot)$ to determine whether $e_i$ appears in the prediction set $\hat{y}_j$:
\begin{equation}
    r_i = M(e_i, \hat{y}_j) \in \{0, 1\},
\end{equation}
where $r_i = 1$ indicates a successful match and $r_i = 0$ otherwise.

The Element Hit Rate for video $j$ is computed as:
\begin{equation}
    \text{EHR}_j = \frac{1}{m} \sum_{i=1}^{m} r_i,
\end{equation}
which represents the recall of ground-truth elements in the prediction.

Finally, the average EHR across all $t$ videos in the dataset is:
\begin{equation}
    \text{Avg.EHR} = \frac{1}{t} \sum_{j=1}^{t} \text{EHR}_j.
\end{equation}
\textbf{Avg.ENHR (Entity Hit Rate):} 
ENHR evaluates whether the reasoning output contains all key entities from the ground-truth label. 
For a given video $j$, let the reasoning output be $\hat{y}_j$ (a free-form text) and the ground-truth entity set be $y_j = \{e_1, e_2, \ldots, e_m\}$, where $m$ denotes the number of ground-truth entities.

To assess whether each ground-truth entity $e_i$ is mentioned in the reasoning output, we employ GPT-4 as a semantic matching function $M(\cdot, \cdot)$:
\begin{equation}
    r_i = M(e_i, \hat{y}_j) \in \{0, 1\},
\end{equation}
where $r_i = 1$ if entity $e_i$ is identified in $\hat{y}_j$, and $r_i = 0$ otherwise.

The Entity Hit Rate for video $j$ is computed as:
\begin{equation}
    \text{ENHR}_j = \frac{1}{m} \sum_{i=1}^{m} r_i,
\end{equation}
which measures the coverage of ground-truth entities in the reasoning output.

The average ENHR across all $t$ videos is:
\begin{equation}
    \text{Avg.ENHR} = \frac{1}{t} \sum_{j=1}^{t} \text{ENHR}_j.
\end{equation}
\textbf{Avg.KHR (Knowledge Hit Rate):} 
KHR evaluates whether the reasoning output incorporates all necessary knowledge points from the ground-truth label. 
For a given video $j$, let the reasoning output be $\hat{y}_j$ (a free-form text) and the ground-truth knowledge set be $y_j = \{k_1, k_2, \ldots, k_m\}$, where $m$ denotes the number of ground-truth knowledge points.

To assess whether each ground-truth knowledge point $k_i$ is utilized in the reasoning output, we employ GPT-4 as a semantic matching function $M(\cdot, \cdot)$:
\begin{equation}
    r_i = M(k_i, \hat{y}_j) \in \{0, 1\},
\end{equation}
where $r_i = 1$ if knowledge point $k_i$ is identified in $\hat{y}_j$, and $r_i = 0$ otherwise.

The Knowledge Hit Rate for video $j$ is computed as:
\begin{equation}
    \text{KHR}_j = \frac{1}{m} \sum_{i=1}^{m} r_i,
\end{equation}
which measures the coverage of ground-truth knowledge in the reasoning output.

The average KHR across all $t$ videos is:
\begin{equation}
    \text{Avg.KHR} = \frac{1}{t} \sum_{j=1}^{t} \text{KHR}_j.
\end{equation}
\textbf{Avg.RHR (Rationale Hit Rate):} 
RHR evaluates whether the reasoning output includes all critical rationales from the ground-truth label. 
For a given video $j$, let the reasoning output be $\hat{y}_j$ (a free-form text) and the ground-truth rationale set be $y_j = \{a_1, a_2, \ldots, a_m\}$, where $m$ denotes the number of ground-truth rationales.

To assess whether each ground-truth rationale $a_i$ is present in the reasoning output, we employ GPT-4 as a semantic matching function $M(\cdot, \cdot)$:
\begin{equation}
    r_i = M(a_i, \hat{y}_j) \in \{0, 1\},
\end{equation}
where $r_i = 1$ if rationale $a_i$ is identified in $\hat{y}_j$, and $r_i = 0$ otherwise.

The Rationale Hit Rate for video $j$ is computed as:
\begin{equation}
    \text{RHR}_j = \frac{1}{m} \sum_{i=1}^{m} r_i,
\end{equation}
which measures the coverage of ground-truth rationales in the reasoning output.

The average RHR across all $t$ videos is:
\begin{equation}
    \text{Avg.RHR} = \frac{1}{t} \sum_{j=1}^{t} \text{RHR}_j.
\end{equation}
\textbf{Avg.AD (Absolute Distance):} 
AD measures the accuracy of shooting angle counting by computing the absolute difference between predicted and ground-truth counts.
For a given video $j$, let $\hat{d}_j$ denote the number of shooting angles predicted by the MLLM and $d_j$ denote the ground-truth count.

The Absolute Distance for video $j$ is computed as:
\begin{equation}
    \text{AD}_j = |\hat{d}_j - d_j|,
\end{equation}
which quantifies the prediction error for that video.

The average AD across all $t$ videos is:
\begin{equation}
    \text{Avg.AD} = \frac{1}{t} \sum_{j=1}^{t} \text{AD}_j.
\end{equation}

A lower Avg.AD indicates better performance in shooting angle counting.
\textbf{Avg.IoU (Intersection Over Union):} 
IoU measures the temporal localization accuracy of key elements by computing the overlap between predicted and ground-truth time ranges.
For a given video $j$, let the predicted time range set be $\hat{y}_j = \{\hat{t}_1, \hat{t}_2, \ldots, \hat{t}_n\}$ and the ground-truth time range set be $y_j = \{t_1, t_2, \ldots, t_m\}$, where $n$ and $m$ denote the number of predicted and ground-truth time ranges, respectively.

For each ground-truth time range $t_i$ in video $j$, we compute its IoU with the best-matching predicted time range:
\begin{equation}
    \text{IoU}_{j,i} = \max_{\hat{t}_k \in \hat{y}_j} \frac{|\hat{t}_k \cap t_i|}{|\hat{t}_k \cup t_i|},
\end{equation}
where $|\hat{t}_k \cap t_i|$ denotes the temporal overlap and $|\hat{t}_k \cup t_i|$ denotes the temporal union between the two ranges.

The average IoU for video $j$ across all its ground-truth time ranges is:
\begin{equation}
    \text{IoU}_j = \frac{1}{m} \sum_{i=1}^{m} \text{IoU}_{j,i}.
\end{equation}

Finally, the average IoU across all $t$ videos is:
\begin{equation}
    \text{Avg.IoU} = \frac{1}{t} \sum_{j=1}^{t} \text{IoU}_j.
\end{equation}
\textbf{Avg. FC (Factual Consistency):} This metric evaluates whether MLLMs' outputs in the NEU task align with ground truth descriptions, quantifying hallucination levels when understanding the overall semantic information of news videos and preventing MLLMs from drawing false conclusions due to hallucinated content. For a given video $j$, let the prediction be $\hat{y}_j$ and the theme understanding ground-truth be $y_j$. We employ GPT-4 as a semantic matching function $M(\cdot, \cdot)$ to quantify the factual consistency between predictions and labels, the prompt as shown in Figure \ref{prompts_for_vfnd_7}. FC scores range from 0 to 5:
\begin{equation}
\text{FC}_j = M(\hat{y}_j, y_j) \in [0, 5],
\end{equation}
where $\text{FC}_j = 5$ indicates perfect consistency and $\text{FC}_j = 0$ indicates complete inconsistency.

Finally, the average FC across all $t$ videos in the dataset is:
\begin{equation}
\text{Avg.FC} = \frac{1}{t} \sum_{j=1}^{t} \text{FC}_j.
\end{equation}
\\
\textbf{Avg. TR (Theme Relevance):}  
This metric evaluates whether MLLMs' outputs in the NEU task generate effective understanding content, avoiding excessive redundant information that may impair the model's reasoning efficiency. For a given video $j$, let the prediction be $\hat{y}_j$ and the theme understanding ground-truth be $y_j$. We employ GPT-4 as a semantic matching function $M(\cdot, \cdot)$ to quantify the theme relevance between predictions and labels, the prompt as shown in Figure \ref{prompts_for_vfnd_7}. TR scores range from 0 to 5:
\begin{equation}
\text{TR}_j = M(\hat{y}_j, y_j) \in [0, 5],
\end{equation}
where $\text{TR}_j = 5$ indicates perfect consistency and $\text{TR}_j = 0$ indicates complete inconsistency.

Finally, the average TR across all $t$ videos in the dataset is:
\begin{equation}
\text{Avg.TR} = \frac{1}{t} \sum_{j=1}^{t} \text{TR}_j.
\end{equation}
\\
\textbf{Avg. CO (Completeness):} This metric evaluates whether MLLMs' outputs in the NEU task generate complete understanding content, avoiding the omission of critical information that could compromise final classification results. For a given video $j$, let the prediction be $\hat{y}_j$ and the theme understanding ground-truth be $y_j$. We employ GPT-4 as a semantic matching function $M(\cdot, \cdot)$ to quantify the factual consistency between predictions and labels, the prompt as shown in Figure \ref{prompts_for_vfnd_7}. FC scores range from 0 to 5:
\begin{equation}
\text{CO}_j = M(\hat{y}_j, y_j) \in [0, 5],
\end{equation}
where $\text{CO}_j = 5$ indicates perfect consistency and $\text{CO}_j = 0$ indicates complete inconsistency.

Finally, the average CO across all $t$ videos in the dataset is:
\begin{equation}
\text{Avg.CO} = \frac{1}{t} \sum_{j=1}^{t} \text{CO}_j.
\end{equation}
\\

\begin{table*}[t]
\centering
\small
\begin{tabular}{p{0.95\textwidth}}
\toprule
\textbf{Prompt:} You are an experienced video fake news detection expert with extensive expertise in identifying video features, video news creation, relevant knowledge and veracity reasoning rationales.
Based on the provided content, please reason and identify the veracity of the video news with the following steps: \\
\textbf{1. If the news is knowledge-oriented,} you need to retrieve general knowledge and the formal content style for creating news of this type, and use them to identify the veracity of the conclusions step by step until obtain the result real or fake. But you cannot directly retrieve or use information on whether the news event occurred before. You cannot retrieve relevant facts that have already occurred based on the time mentioned in the news. \\[0.5em]
\textbf{2. If the news is content-oriented,} please reason about its veracity according to the following steps until obtain the result real or fake: \\
\quad (1) Distinguish between original shot footage and creator-added content... \\
\quad (2) Identify creator-added text details: text content, color, position on the screen. \\
\quad (3) Identify key visual elements critical to authenticity assessment... \\
\quad (4) Temporal grounding and detailed analysis of key elements, locate the time segments... \\
\quad (5) Identify all individuals appearing in the video. Determine the identities/roles of all people shown, such as: parties involved in the incident, law enforcement officers, victims, medical personnel, other relevant persons. \\
\quad (6) Count shooting angles. \\[0.5em]
\textbf{3. If the news is knowledge and content oriented}, please verify the veracity of event in the video based on the step 1 and 2. \\[0.5em]
\textbf{4. Final output:} Based on the results of step 1, 2, 3 and your professional knowledge, reason about the veracity of this news video and output the result (fake/real) with the following format: \\
If the video news is more likely to be real, output \texttt{<think>your reasoning process</think><result>real</result>}; \\
Otherwise, output \texttt{<think>your reasoning process</think><result>fake</result>}. \\
\bottomrule
\end{tabular}
\caption{Detail of POVFND-CoT.}
\label{prompt4_mvfnd_cot}
\label{tab:prompt}
\end{table*}


\begin{table*}[t]
\centering
\small
\begin{tabular}{p{0.95\textwidth}}
\toprule
\textbf{Prompt:} You are an experienced video fake news detection expert with extensive expertise in identifying video features, video news creation, relevant knowledge and veracity reasoning rationales. The video news title is '\{text\}'. The associated speech transcript for the video is '\{transcript\}'. The creator-added content in the video are '\{lm\_content\}'. The descriptions for creator-added content font colors in the video are '\{lm\_color\}'. The descriptions for creator-added content positions in the video are '\{lm\_position\}'. The key elements in the video are '\{lm\_shooting\_key\_elements\}'. The identities of the characters appearing in the video are '\{lm\_shooting\_role\}'. The shooting angle in the video is '\{lm\_shooting\_key\_shots\}'. Based on the provided content, please reason and identify the veracity of the video news with the following steps: \\[0.5em]
\textbf{1. If the news is knowledge-oriented,} you need to retrieve general knowledge and the formal content style for creating news of this type, and use them to identify the veracity of the conclusions step by step until obtain the result real or fake. But you cannot directly retrieve or use information on whether the news event occurred before. You cannot retrieve relevant facts that have already occurred based on the time mentioned in the news. \\[0.5em]
\textbf{2. If the news is content-oriented,} please reason about its veracity according to the following steps until obtain the result real or fake: \\
\quad (1) Distinguish between original shot footage and creator-added content. Identify and separate the authentic filmed content from elements added by the video creator during post-production. \\
\quad (2) Identify creator-added text details. Detect all text overlays added by the creator, including: the specific text content, text color, text position on the screen (e.g., top-left, center, bottom-right). \\
\quad (3) Identify key visual elements critical to authenticity assessment. Based on your expertise, identify key frames/elements that significantly impact the video's credibility, including but not limited to: on-scene footage of the reported event, clear facial shots of individuals, official announcements or notices, other relevant visual evidence. \\
\quad (4) Temporal grounding and detailed analysis of key elements, locate the time segments where these key visual elements appear in the video, conduct a more detailed examination of these segments, extract clues that could help verify the authenticity of the news content. \\
\quad (5) Identify all individuals appearing in the video. Determine the identities/roles of all people shown, such as: parties involved in the incident, law enforcement officers, victims, medical personnel, other relevant persons. \\
\quad (6) Count shooting angles. Determine the total number of distinct camera angles/perspectives used throughout the entire video. \\[0.5em]
\textbf{3. If the news is knowledge and content oriented,} please verify the veracity of event in the video based on the step 1 and 2. \\[0.5em]
\textbf{4. Final output:} Based on the results of step 1, 2, 3 and your professional knowledge, reason about the veracity of this news video step by step and output the result (fake/real) with the following format: \\
If the video news is more likely to be real, output \texttt{<think>your reasoning process</think><result>real</result>}; \\
Otherwise, output \texttt{<think>your reasoning process</think><result>fake</result>}. \\
\bottomrule
\end{tabular}
\caption{Detail of ALL-POVFND-CoT.}
\label{prompt4_all_mvfnd_cot}
\end{table*}

\begin{table*}[t]
\centering
\small
\begin{tabular}{p{0.95\textwidth}}
\toprule
\textbf{Prompt:} You are an experienced video fake news detection expert with extensive expertise in identifying video features, video news creation, relevant knowledge and veracity reasoning rationales. The video news title is `\{text\}'. The associated speech transcript for the video is `\{transcript\}'. The creator-added content in the video are `\{lm\_content\}'. The descriptions for creator-added content font colors in the video are `\{lm\_color\}'. The descriptions for creator-added content positions in the video are `\{lm\_position\}'. Based on the provided content, please reason and identify the veracity of the video news with the following steps: \\[0.5em]
\textbf{1. If the news is knowledge-oriented,} you need to retrieve general knowledge and the formal content style for creating news of this type, and use them to identify the veracity of the conclusions step by step until obtain the result real or fake. But you cannot directly retrieve or use information on whether the news event occurred before. You cannot retrieve relevant facts that have already occurred based on the time mentioned in the news. \\[0.5em]
\textbf{2. If the news is content-oriented,} please reason about its veracity according to the following steps until obtain the result real or fake: \\
\quad (1) Distinguish between original shot footage and creator-added content. Identify and separate the authentic filmed content from elements added by the video creator during post-production. \\
\quad (2) Identify creator-added text details. Detect all text overlays added by the creator, including: the specific text content, text color, text position on the screen (e.g., top-left, center, bottom-right). \\
\quad (3) Identify key visual elements critical to authenticity assessment. Based on your expertise, identify key frames/elements that significantly impact the video's credibility, including but not limited to: on-scene footage of the reported event, clear facial shots of individuals, official announcements or notices, other relevant visual evidence. \\
\quad (4) Temporal grounding and detailed analysis of key elements, locate the time segments where these key visual elements appear in the video, conduct a more detailed examination of these segments, extract clues that could help verify the authenticity of the news content. \\
\quad (5) Identify all individuals appearing in the video. Determine the identities/roles of all people shown, such as: parties involved in the incident, law enforcement officers, victims, medical personnel, other relevant persons. \\
\quad (6) Count shooting angles. Determine the total number of distinct camera angles/perspectives used throughout the entire video. \\[0.5em]
\textbf{3. If the news is `knowledge and content oriented',} please verify the veracity of event in the video based on the step 1 and 2. \\[0.5em]
\textbf{4. Final output:} Based on the results of step 1, 2, 3 and your professional knowledge, reason about the veracity of this news video step by step and output the result (fake/real) with the following format: If the video news is more likely to be real, output \texttt{<think>your reasoning process</think><result>real</result>}; Otherwise, output \texttt{<think>your reasoning process</think><result>fake</result>}. Once you determine the detection conclusion, you should immediately terminate the reasoning process and output the result in the above format. \\
\bottomrule
\end{tabular}
\caption{Detail of CAC-POVFND-CoT.}
\label{prompt4_cac_mvfnd_cot}
\end{table*}

\begin{table*}[t]
\centering
\small
\begin{tabular}{p{0.95\textwidth}}
\toprule
\textbf{Prompt:} You are an experienced video fake news detection expert with extensive expertise in identifying video features, video news creation, relevant knowledge and veracity reasoning rationales. The video news title is `\{text\}'. The associated speech transcript for the video is `\{transcript\}'. The key elements in the video are `\{lm\_shooting\_key\_elements\}'. The key elements time range in the video are `\{lm\_shooting\_key\_elements\_temporal\_grounding\}'. The identities of the characters appearing in the video are `\{lm\_shooting\_role\}'. The shooting angle in the video is `\{lm\_shooting\_key\_shots\}'. Based on the provided content, please reason and identify the veracity of the video news with the following steps: \\[0.5em]
\textbf{1. If the news is knowledge-oriented,} you need to retrieve general knowledge and the formal content style for creating news of this type, and use them to identify the veracity of the conclusions step by step until obtain the result real or fake. But you cannot directly retrieve or use information on whether the news event occurred before. You cannot retrieve relevant facts that have already occurred based on the time mentioned in the news. \\[0.5em]
\textbf{2. If the news is content-oriented,} please reason about its veracity according to the following steps until obtain the result real or fake: \\
\quad (1) Distinguish between original shot footage and creator-added content. Identify and separate the authentic filmed content from elements added by the video creator during post-production. \\
\quad (2) Identify creator-added text details. Detect all text overlays added by the creator, including: the specific text content, text color, text position on the screen (e.g., top-left, center, bottom-right). \\
\quad (3) Identify key visual elements critical to authenticity assessment. Based on your expertise, identify key frames/elements that significantly impact the video's credibility, including but not limited to: on-scene footage of the reported event, clear facial shots of individuals, official announcements or notices, other relevant visual evidence. \\
\quad (4) Temporal grounding and detailed analysis of key elements, locate the time segments where these key visual elements appear in the video, conduct a more detailed examination of these segments, extract clues that could help verify the authenticity of the news content. \\
\quad (5) Identify all individuals appearing in the video. Determine the identities/roles of all people shown, such as: parties involved in the incident, law enforcement officers, victims, medical personnel, other relevant persons. \\
\quad (6) Count shooting angles. Determine the total number of distinct camera angles/perspectives used throughout the entire video. \\[0.5em]
\textbf{3. If the news is `knowledge and content oriented',} please verify the veracity of event in the video based on the step 1 and 2. \\[0.5em]
\textbf{4. Final output:} Based on the results of step 1, 2, 3 and your professional knowledge, reason about the veracity of this news video step by step and output the result (fake/real) with the following format: If the video news is more likely to be real, output \texttt{<think>your reasoning process</think><result>real</result>}; Otherwise, output \texttt{<think>your reasoning process</think><result>fake</result>}. Once you determine the detection conclusion, you should immediately terminate the reasoning process and output the result in the above format. \\
\bottomrule
\end{tabular}
\caption{Detail of OSF-POVFND-CoT.}
\label{prompt4_osf_mvfnd_cot}
\end{table*}

\section{Evaluation Prompts}

\subsection{Task Evaluation Prompts}
\begin{figure*}[t]
  \centering
  \includegraphics[width=\textwidth]{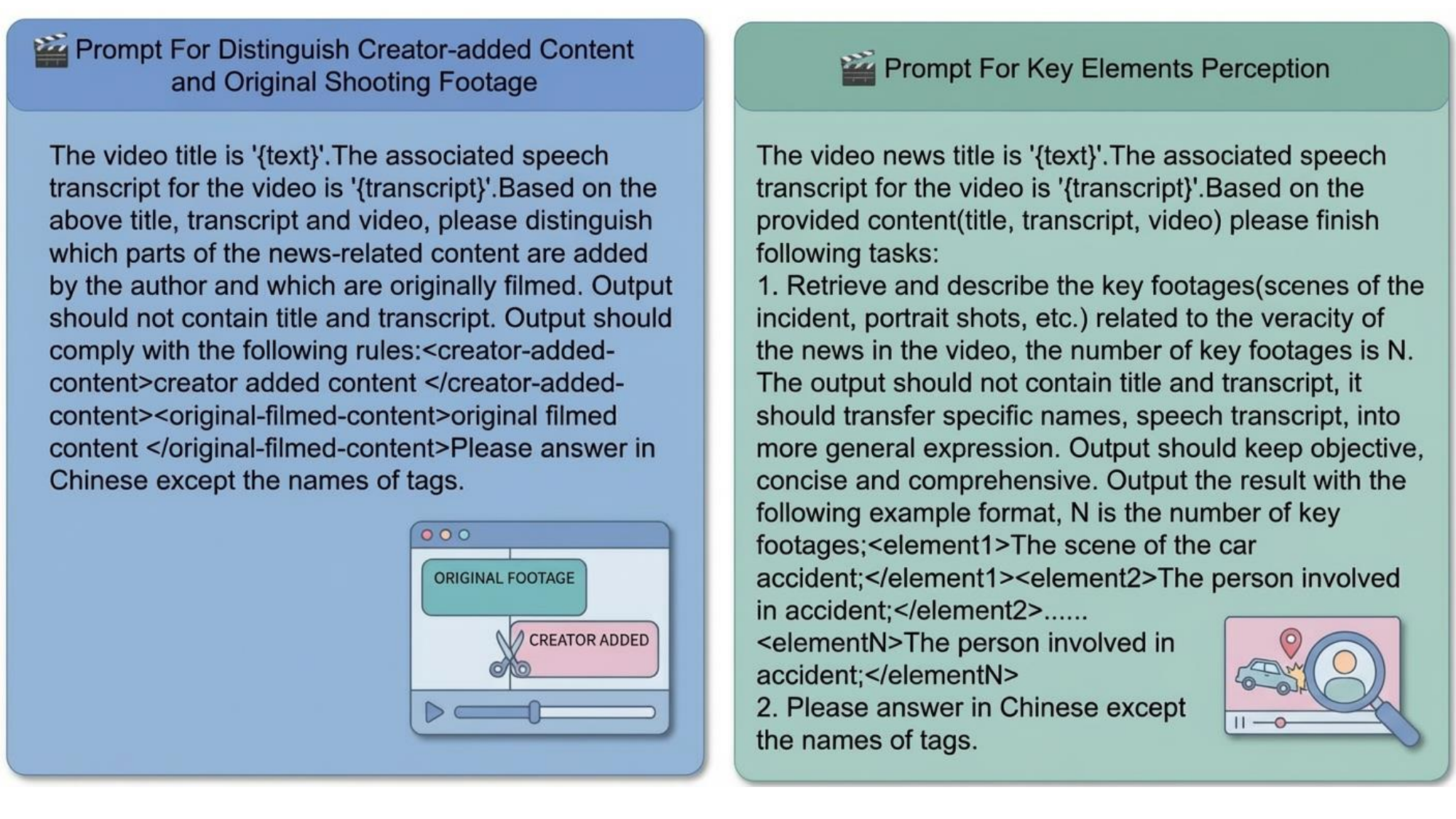}  
  \captionsetup{justification=centering, singlelinecheck=false}
  \caption{Prompt For DCS and KEP.}
  \label{prompts_for_vfnd_1}
\end{figure*}

\begin{figure*}[t]
  \centering
  \includegraphics[width=\textwidth]{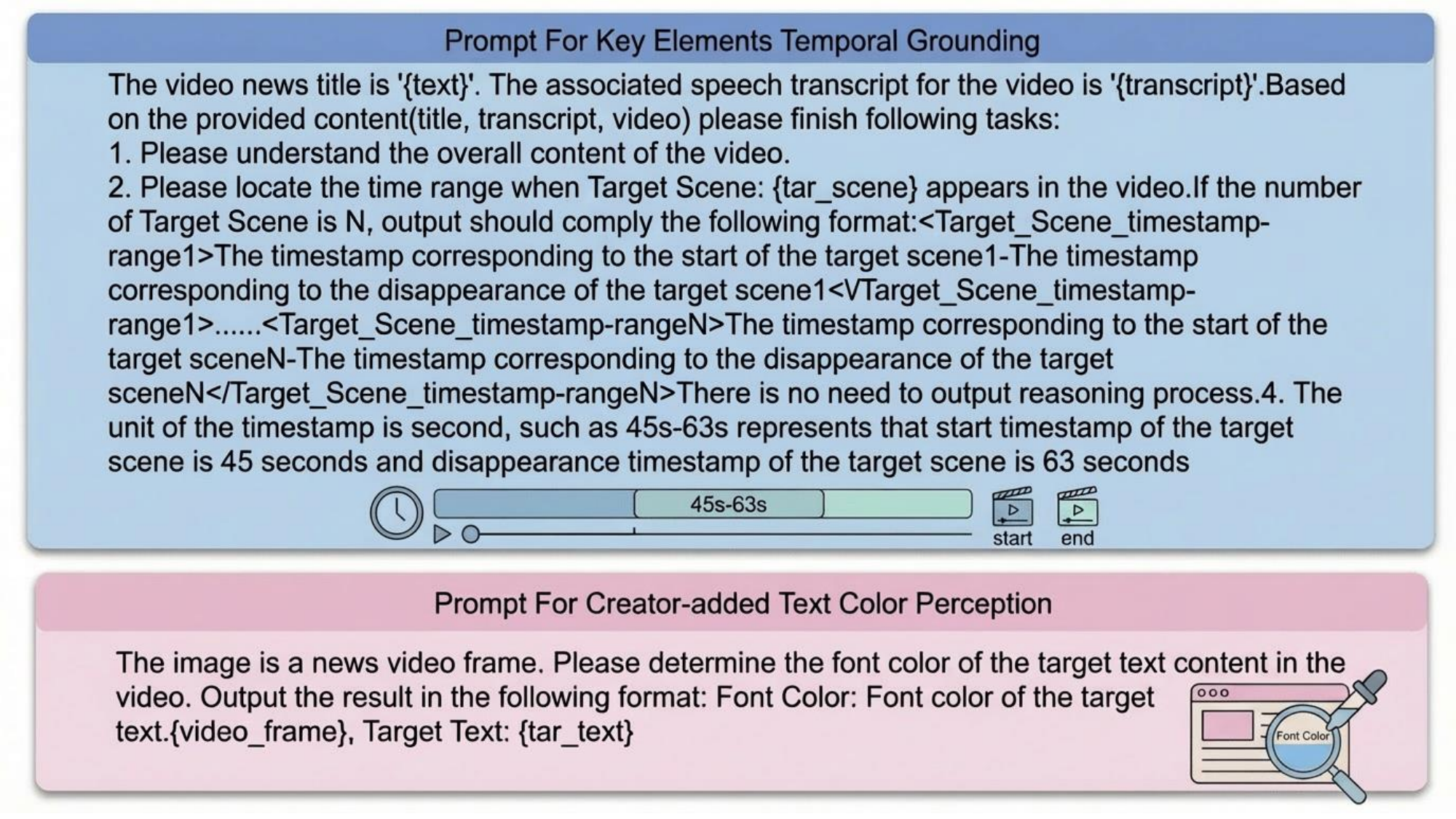}  
  \captionsetup{justification=centering, singlelinecheck=false}
  \caption{Prompt For KEG and CCP.}
  \label{prompts_for_vfnd_2}
\end{figure*}

\begin{figure*}[t]
  \centering
  \includegraphics[width=\textwidth]{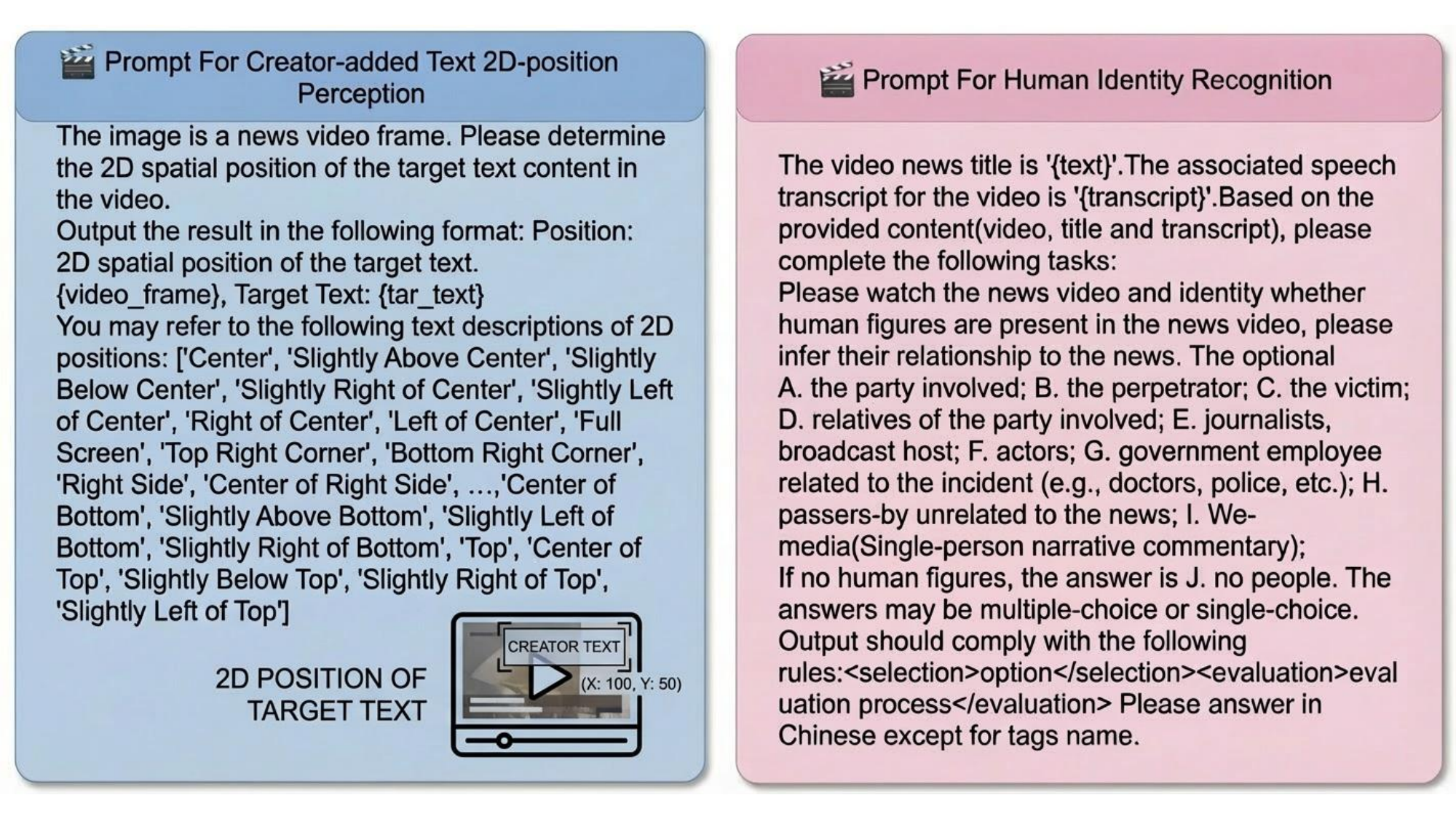}  
  \captionsetup{justification=centering, singlelinecheck=false}
  \caption{Prompt For CPP and HIR.}
  \label{prompts_for_vfnd_3}
\end{figure*}

\begin{figure*}[t]
  \centering
  \includegraphics[width=\textwidth]{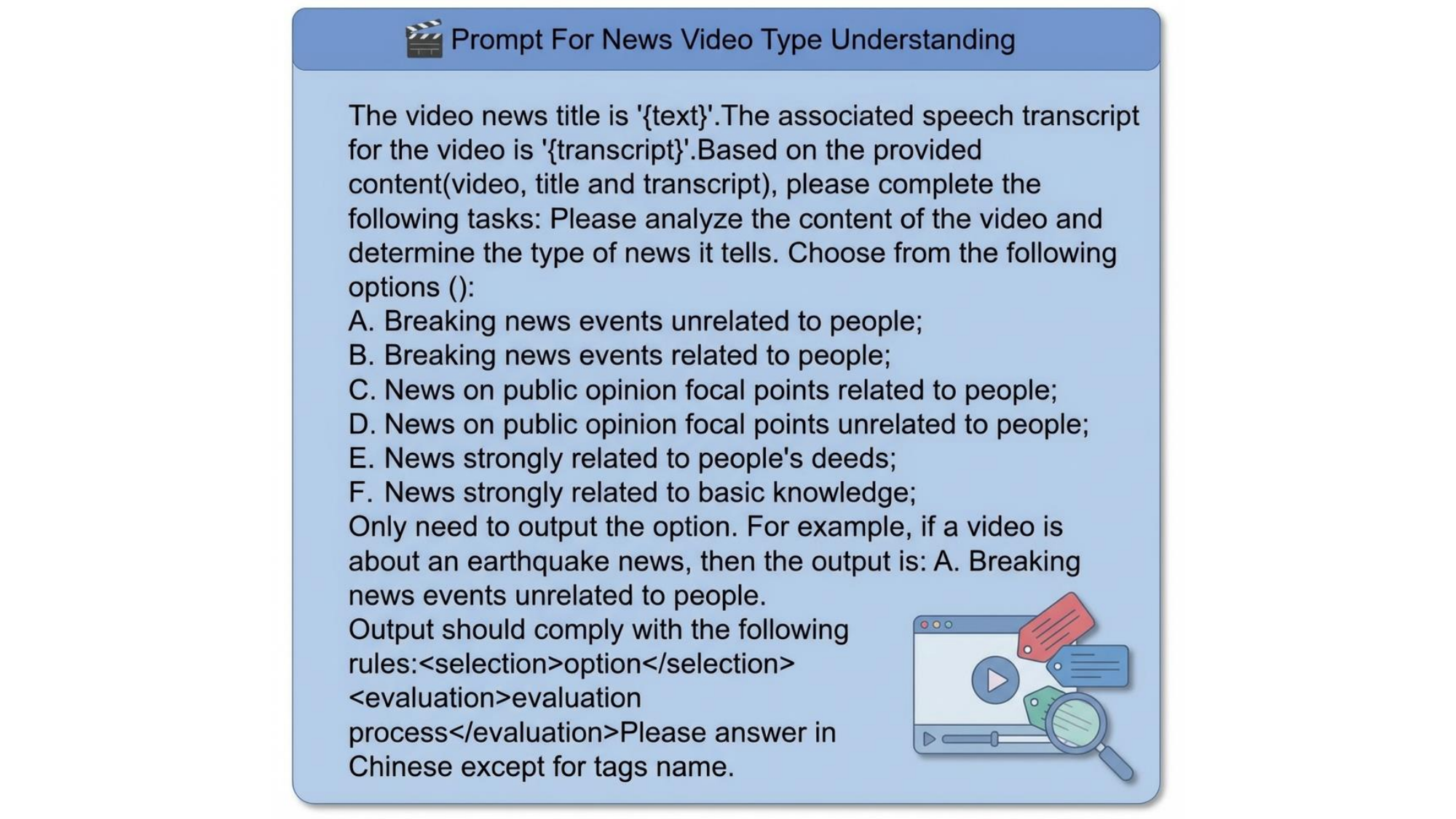}  
  \captionsetup{justification=centering, singlelinecheck=false}
  \caption{Prompt For NTU.}
  \label{prompts_for_vfnd_4}
\end{figure*}

\begin{figure*}[t]
  \centering
  \includegraphics[width=\textwidth]{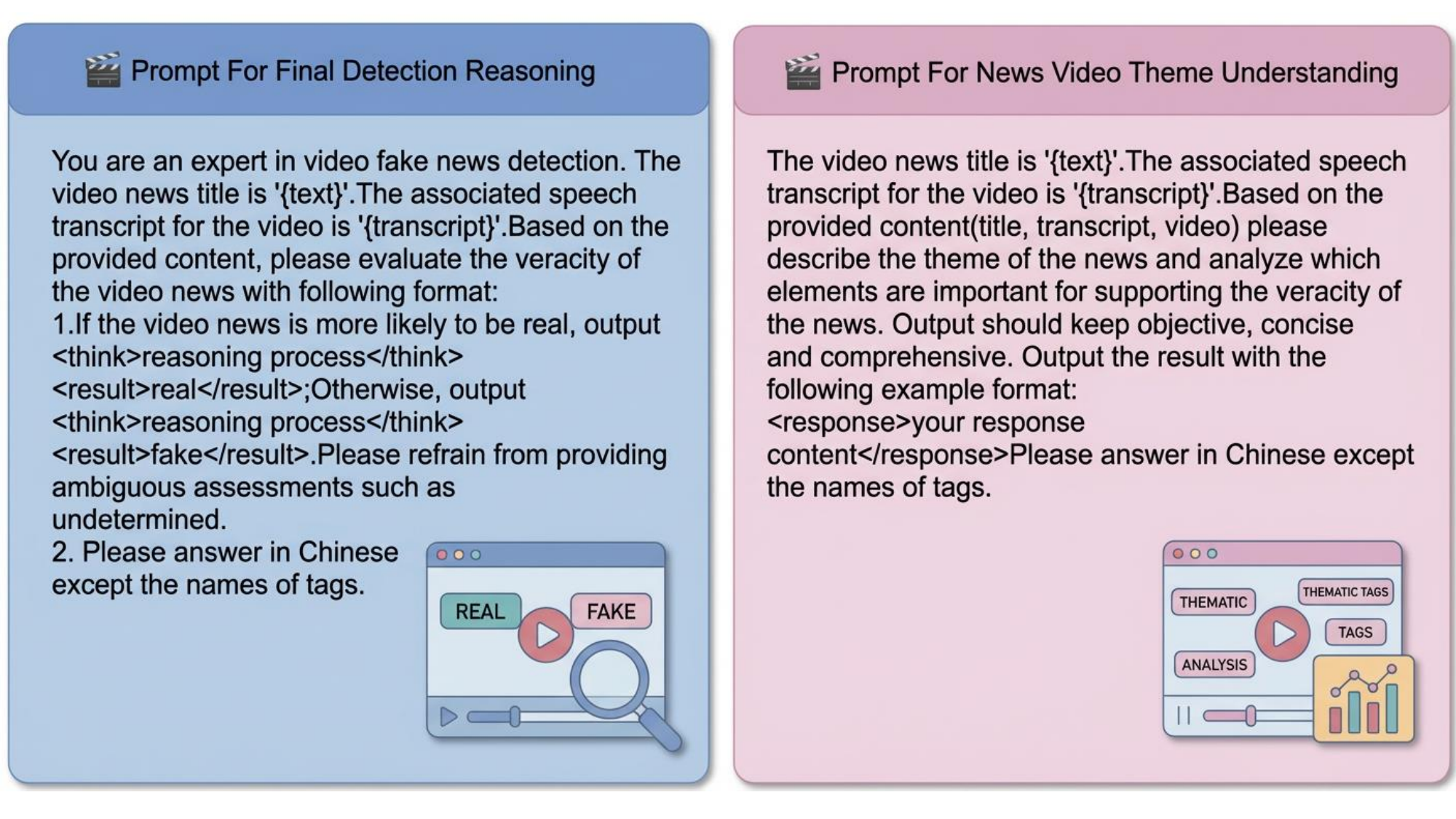}  
  \captionsetup{justification=centering, singlelinecheck=false}
  \caption{Prompt For FDR and NEU.}
  \label{prompts_for_vfnd_5}
\end{figure*}

\begin{figure*}[t]
  \centering
  \includegraphics[width=\textwidth]{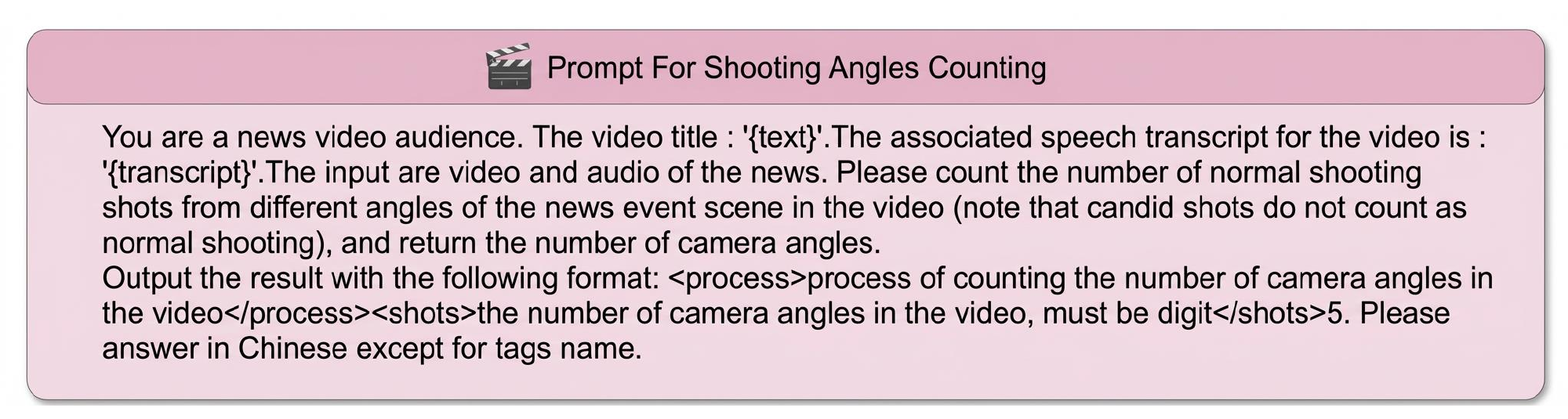}  
  \captionsetup{justification=centering, singlelinecheck=false}
  \caption{Prompt For SAC}
  \label{prompts_for_vfnd_6}
\end{figure*}

To provide detailed evaluation procedures, each task is evaluated under a zero-shot prompting paradigm. MLLMs are instructed to generate outputs in specified formats, with specific prompts shown in Figures \ref{prompts_for_vfnd_1}--\ref{prompts_for_vfnd_6}. Our tasks encompass multiple response formats including open-ended questions, single-choice, multiple-choice, and structured outputs. For evaluation efficiency, we incorporate relevant contextual information into the prompts. For instance, to accurately assess MLLMs' spatial perception capacity in the CPP task, we provide pre-defined position descriptions covering the screen, requiring models to select the appropriate semantic representation from the given options (see Figure \ref{prompts_for_vfnd_3} for details).

\subsection{Results Evaluation Prompts}
\begin{figure*}[t]
  \centering
  \includegraphics[width=\textwidth]{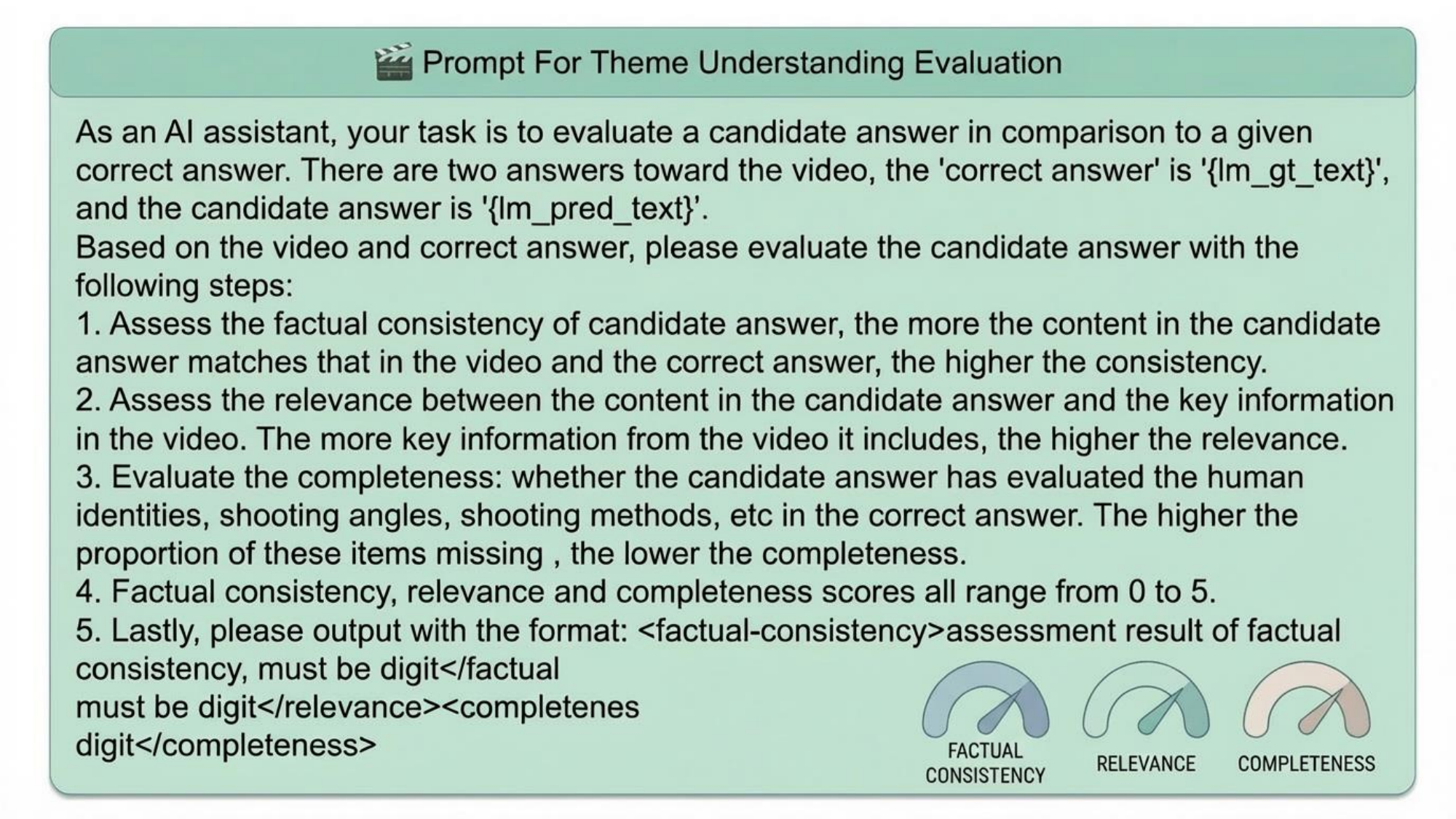}  
  \captionsetup{justification=centering, singlelinecheck=false}
  \caption{Prompt For NEU Outputs Evaluation.}
  \label{prompts_for_vfnd_7}
\end{figure*}

\begin{figure*}[t]
  \centering
  \includegraphics[width=\textwidth]{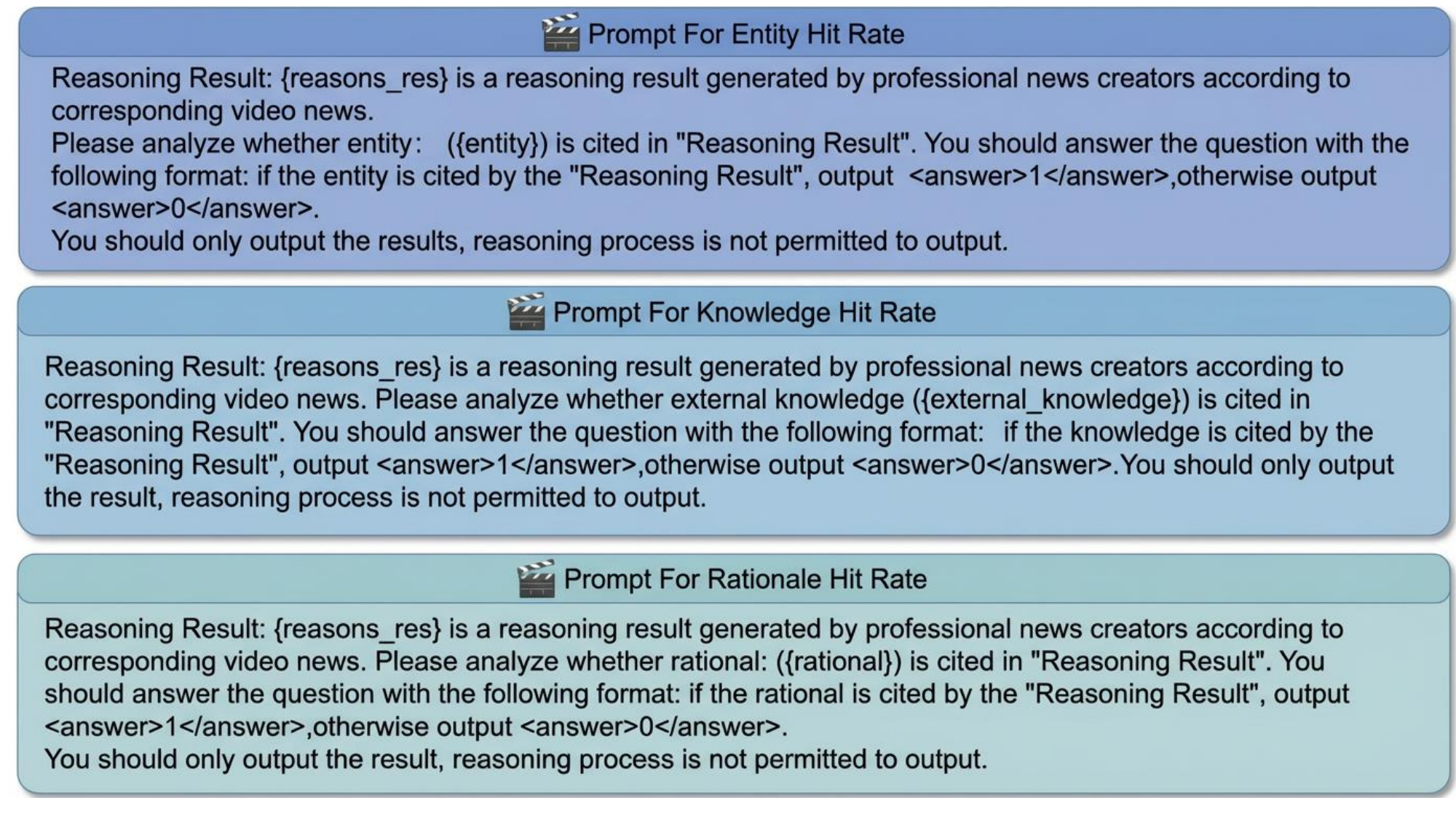}  
  \captionsetup{justification=centering, singlelinecheck=false}
  \caption{Prompt For FDR Outputs Evaluation.}
  \label{prompts_for_vfnd_8}
\end{figure*}

In addition to exact match metrics, we introduce semantic match metrics—including factual consistency, theme relevance, completeness and elements/entity/knowledge/rationale hit rate to evaluate open-ended output tasks (KEP, NEU and FDR). We employ GPT-4o as an evaluator, using prompts shown in Figures \ref{prompts_for_vfnd_7} and \ref{prompts_for_vfnd_8} to assess MLLMs' key elements perception, theme understanding and veracity reasoning performance.

\subsection{Error Case Analysis}
To investigate the actual impact of different visual prompts on MVFNC-CoT results, we conduct the following case study. We find that CAC and OSF provide distinct benefits to CoT, as illustrated in Figure \ref{error_analysis_frames_1} and Figure \ref{error_analysis_frames_2}, respectively.
\begin{table*}[!t]
\centering
\small
\setlength{\tabcolsep}{4pt}
\begin{tabular}{>{\centering\arraybackslash}m{2.8cm}m{12.7cm}}
\toprule
\textbf{Micro-video} & 
\includegraphics[width=12.5cm,height=3cm]{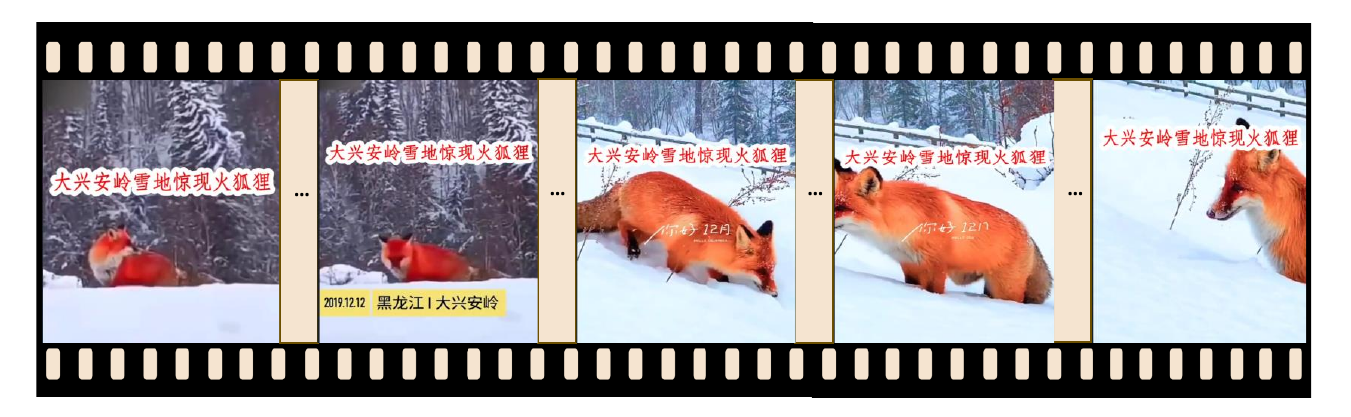} \\
\midrule
\textbf{Title} & 
Fire Fox Surprisingly Appeared in the Snow of Greater Khingan Mountains \\
\midrule
\multicolumn{2}{c}{
\begin{tabular}[t]{@{}>{\centering\arraybackslash}p{7.75cm}|>{\centering\arraybackslash}p{7.75cm}@{}}
\textbf{POVFND-CoT (Incorrect)} & \textbf{CAC-POVFND-CoT (Correct)} \\
\hline
\raggedright ... discovery of a red fox in the snow of Greater Khingan Mountains. Retrieval verified that red foxes (Vulpes vulpes) are widely distributed in China, especially in northern forest regions, and \textcolor{red}{Greater Khingan Mountains is within their known habitat. Therefore, discovering red foxes in this area is biologically reasonable.} The video shows multiple clips of the fox in the snow, exhibiting natural behaviors such as resting and foraging, consistent with wild fox habits. The footage appears natural without obvious staging or post-production manipulation. The fox's fire-red color matches typical characteristics of red foxes. The snow and forest background in the video matches the winter environment of Greater Khingan Mountains...\textcolor{red}{The color and position of the text also conform to the presentation style of general news content, without the use of excessively exaggerated or highly emotionally suggestive styles.}<result>\textcolor{red}{Real}</result> & 
\raggedright ..., and the Greater Khingan Range is precisely among red foxes' known habitats. Therefore, the discovery of red foxes in the snowy landscapes of the Greater Khingan Range is entirely biologically plausible. However, judging from the content of the news video,... \textcolor{blue}{The added text "Fire Fox Surprisingly Appeared" uses the emotionally charged word "surprisingly" with red font color, which contradicts the objective, neutral, and professional style in news titles. Moreover, the text is placed in the center of the screen, obscuring part of the original footage, and such unprofessional editing methods have greatly reduced its authenticity. }.... It mentions "whether it is the character Fengjiu from the TV series," positioning it as entertainment content rather than serious science education, failing to meet the depth and professionalism standards of knowledge-oriented news. <result>\textcolor{blue}{Fake}</result> \\
\end{tabular}} \\
\midrule
\textbf{Ground Truth} & \textcolor{blue}{Fake} \\
\bottomrule
\end{tabular}
\caption{Error case analysis demonstrating the difference between POVFND-CoT and CAC-POVFND-CoT reasoning processes in VFND.}
\label{error_analysis_frames_1}
\end{table*}

\begin{table*}[!t]
\centering
\small
\setlength{\tabcolsep}{4pt}
\begin{tabular}{>{\centering\arraybackslash}m{2.8cm}m{12.7cm}}
\toprule
\textbf{Micro-video} & 
\includegraphics[width=12.5cm,height=3cm]{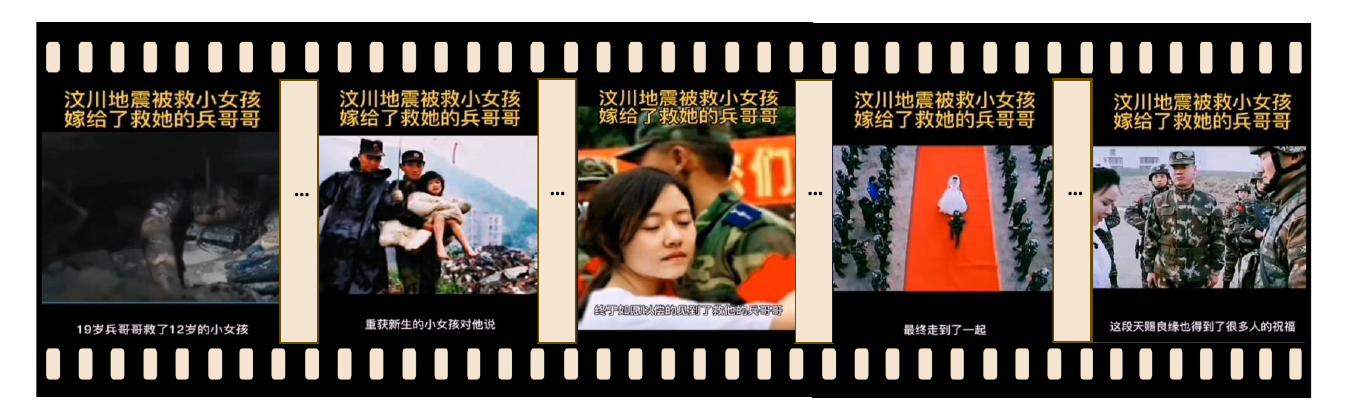} \\
\midrule
\textbf{Title} & 
Girl Rescued in Wenchuan Earthquake Married Her Rescuer Soldier \\
\midrule
\multicolumn{2}{c}{
\begin{tabular}[t]{@{}>{\centering\arraybackslash}p{7.75cm}|>{\centering\arraybackslash}p{7.75cm}@{}}
\textbf{POVFND-CoT (Incorrect)} & \textbf{CAC-POVFND-CoT (Correct)} \\
\hline
\raggedright \textcolor{red}{The Wenchuan earthquake mentioned in the video did occur in 2008, and 13 years have passed since then, which matches the timeline.} ... The wedding scenes and rescue scenes shown in the video, as well as the characters' clothing and backgrounds, correspond to the described plot, adding authenticity to the story. ... The character dialogues and subtitles in the video detail the entire story development, making it appear coherent and logical. ... Although this coincidental and romantic story sounds somewhat dramatic, its authenticity cannot be completely ruled out, especially since deep emotional connections between people during major disasters are not uncommon. Meanwhile, \textcolor{red}{the text "Girl Rescued in Wenchuan Earthquake Married Her Rescuer Soldier" at the top center of the video is in yellow font, which in terms of text placement and font color appears to be characteristic of authentic news. Based on the above analysis, the news video is more likely to be real.} <result>\textcolor{red}{Real}</result> & 
\raggedright ... \textcolor{blue}{The footage used in the video, such as the girl under earthquake rubble, old photos of soldiers carrying the girl, reunion scenes, and wedding scenes, may come from different times and events. Moreover, the opening footage is from the movie "Aftershock," lacking the documentary feel of authentic news reporting.} This splicing of footage from various sources without clear attribution undermines the news's credibility. \textcolor{blue}{Additionally, the news is mainly narrated by a self-media personality with strong emotional bias, appearing more like attention-seeking content rather than news reporting, greatly reducing its authenticity.} Although the Wenchuan earthquake is a real event, the video combines earthquake rescue with a dramatic "grown-up marrying her savior" love story, claiming this promise was ultimately fulfilled. Such highly emotional, legendarily-colored narratives, without detailed reports and verification from authoritative media, ...<result>\textcolor{blue}{Fake}</result> \\
\end{tabular}} \\
\midrule
\textbf{Ground Truth} & \textcolor{blue}{Fake} \\
\bottomrule
\end{tabular}
\caption{Error case analysis demonstrating the difference between POVFND-CoT and CAC-POVFND-CoT reasoning processes in VFND.}
\label{error_analysis_frames_2}
\end{table*}

\section{Answers For Submission}
\subsection{Potential Risks}
\hspace*{1em} \textbf{1) Ignoring Adversarial Evolution of Fake Videos}. A benchmark’s value lies in its ability to reflect real-world deployment challenges. We may reduce our work’s relevance if we overlook these risks, fake news creators continuously adapt to detection tools—for example, using adversarial deepfakes (designed to evade LLMs), mixing fake and real footage, or leveraging AI to generate contextually consistent manipulations. If our benchmark uses static, outdated fake samples :1. Our results will quickly become obsolete as new manipulation techniques emerge.2. The benchmark will not test models’ ability to generalize to "future" fake video types, limiting its long-term utility for industry or policy.\\
\hspace*{1em} \textbf{2) Lack of Cross-Domain Validation}. Fake news videos appear in diverse domains (e.g., politics, healthcare, entertainment), each with unique characteristics (e.g., medical deepfakes use specialized terminology; political fakes rely on event context). If our benchmark focuses on a single domain: 1. Models optimized for our benchmark may fail in critical other domains (e.g., a model good at detecting political fakes may miss fake medical advice videos). 2. We cannot support cross-domain comparisons, a key requirement for organizations (e.g., health agencies, social media platforms) that need tools for multiple use cases.\\
\hspace*{1em} \textbf{2) Vague or Narrow Definition of "Fake News Video"}. Fake news videos encompass a spectrum of manipulations (e.g., deepfakes, selective editing, context misattribution) and intents (malicious disinformation vs. accidental misinformation). If we use an imprecise or overly narrow definition (e.g., only labeling face swaps as "fake"), we create inconsistency: Models optimized for specific fake types (e.g., deepfake detection) will be unfairly penalized for failing to detect other valid fake cases (e.g., a real video paired with misleading audio).

\subsection{The License For Artifacts}
In our research on multimodal large model-based fake news video detection, we recognize that clarifying the license and usage terms of core artifacts is critical to ensuring academic compliance. For the Fake News Short Video Dataset (FakeSV-2023) — a key resource supporting our benchmark construction — we have engaged in in-depth discussions with the dataset authors and reached a formal agreement on its use and distribution terms. This agreement not only specifies that the dataset is strictly limited to non-commercial research purposes, prohibiting any commercial exploitation such as resale or use in commercial product development, but also mandates mandatory citation of its original AAAI 2023 publication (in the specified BibTeX format) in all our related outputs. Additionally, it outlines clear responsibilities: we and our affiliated institution assume full liability for any consequences of dataset use and must indemnify the authors against claims arising therefrom, while the authors retain the right to terminate our access if necessary. We have formally consented to all these provisions, and the agreement details have been systematically documented in our research records to ensure every step of our work aligns with the dataset’s usage specifications.

\subsection{Intended Use}
Regarding the use of the Fake News Short Video Dataset (FakeSV-2023) in our research, we confirm that our usage fully aligns with the intended use specified in the license agreement reached with the dataset authors. As explicitly stipulated in the agreement, the Dataset is restricted to non-commercial research purposes only—a scope that precisely matches our intended application of the Dataset in constructing and validating the multimodal large model-based fake news video detection benchmark.
Throughout the entire research process, we have strictly adhered to this boundary: the Dataset and any derived data generated from it (e.g., preprocessed video clips, annotated sub-datasets for model training) have been exclusively used for academic research activities, including algorithm development, performance testing, and result analysis related to fake news video detection. No part of the Dataset or its derivatives has been employed in any scenarios outside of research, such as commercial product development, commercial service provision, or any other non-academic uses that are prohibited by the agreement. This compliance ensures that our use of the Dataset remains consistent with both the authors’ intended design for the Dataset and the original access conditions outlined in the license terms.

\subsection{Model Size And Budget}
Regarding the reporting of model parameters, total computational budget, and computing infrastructure used in our research, we provide the following details:
For the state-of-the-art multimodal large language models (LLMs) employed, we distinguish between closed-source and open-source models due to differences in parameter disclosure and deployment methods: 1. Closed-source models: We accessed these via cloud service APIs. Specifically, we made a total of 137,927 API calls to the Gemini 2.5-Flash model, and a cumulative 315,061 API calls to GPT series models (including GPT-4o-, GPT-4o-mini, and GPT-o1). Note that the exact number of parameters for these closed-source models is not publicly disclosed by their developers, so we do not report proprietary parameter details. 2. Open-source models: These were deployed on our in-house computing infrastructure, which consists of 8 H800 GPUs. The total computational budget for training and inference with these open-source multimodal LLMs amounted to approximately 1,500 GPU hours.

\subsection{Recruitment And Payment}
We hired a total of 35 data annotators on a part-time basis, with hourly remuneration ranging from HK\$60 to HK\$100. This payment rate is higher than the local statutory minimum wage. Details of their specific circumstances are provided in Section 4.2.
\subsection{Use Of Ai Assistants}
Assisting with program bug fixes: During the development and debugging of code for model deployment (e.g., integrating open-source multimodal LLMs with our experimental framework) and data preprocessing (e.g., optimizing video frame extraction scripts), we used AI assistants to identify syntax errors, logical inconsistencies, and performance bottlenecks in the code. Their role was limited to providing diagnostic suggestions and potential solution references, with final bug fixes and code validation completed by our research team to ensure accuracy and alignment with our experimental objectives.\\

\end{document}